\documentclass[11pt]{article}
\usepackage{acl}
\usepackage{times}
\usepackage{latexsym}
\usepackage{tcolorbox}
\usepackage{enumitem}
\usepackage[T1]{fontenc}

\usepackage[utf8]{inputenc}
\usepackage{microtype}
\usepackage{inconsolata}
\usepackage{graphicx}
\usepackage{amsmath}
\usepackage{amssymb}
\usepackage{amsfonts}

\usepackage{natbib}

\usepackage{float}
\usepackage{booktabs}
\usepackage{multirow}
\usepackage[dvipsnames]{xcolor}
\usepackage{subcaption}

\title{Abstract Activation Spaces for Content-Invariant Reasoning \\in Large Language Models}

\author{
  Gabriele Maraia $^{(\dagger)}$ Marco Valentino$^{(\bullet)}$  \\
 \textbf{Fabio Massimo Zanzotto} $^{(\dagger,\ddagger )}$ \textbf{Leonardo Ranaldi} $^{(\oplus, \dagger)}$ \\
  ${(\dagger)}$ Human Centric ART, University of Rome Tor Vergata \\
 	${(\oplus)}$ ILCC, School of Informatics, University of Edinburgh \\
     	${(\bullet)}$ School of Computer Science,  University of Sheffield, ${(\ddagger)}$ Almawave S.p.A.  \\
\small  \texttt{\{first\_name.last\_name\}@uniroma2.it}
}

\usepackage[compact]{titlesec}
\titlespacing{\section}{0pt}{5pt}{3pt}
\titlespacing{\subsection}{0pt}{4pt}{2pt}
\titlespacing{\paragraph}{0pt}{2pt}{1em}

\begin{document}
\maketitle
\begin{abstract}
Large Language Models (LLMs) often struggle with deductive judgment in syllogistic reasoning, systematically conflating semantic plausibility with formal validity--a phenomenon known as \emph{content effect}. This bias persists even when models generate step-wise explanations, indicating that intermediate rationales may inherit the same semantic shortcuts that affect answers.
Recent approaches propose mitigating this issue by increasing inference-time structural constraints, either by encouraging abstract intermediate representations or by intervening directly in the model’s internal computations; however, reliably suppressing semantic interference remains an open challenge.

To make formal deduction less sensitive to semantic content, we introduce a framework for abstraction-guided reasoning that explicitly separates structural inference from lexical semantics. We construct paired \emph{content-laden} and \emph{abstract} syllogisms and use the model’s activations on abstract inputs to define an \emph{abstract reasoning space}. We then learn lightweight \emph{Abstractors} that, from content-conditioned residual-stream states, predict representations aligned with this space and integrate these predictions via multi-layer interventions during the forward pass. Using cross-lingual transfer as a test bed, we show that abstraction-aligned steering reduces content-driven errors and improves validity-sensitive performance. 
Our results position activation-level abstraction as a scalable mechanism for enhancing the robustness of formal reasoning in LLMs against semantic interference.
\end{abstract}

\section{Introduction}
LLMs have demonstrated remarkable capabilities across a wide range of complex reasoning tasks. Yet, they frequently privilege semantic intuition over formal logic, systematically struggling to disentangle the \textit{form} of an argument from its \textit{content} \cite{eisape-etal-2024-systematic,dasgupta2024languagemodelshumanlikecontent}. This limitation becomes particularly evident in syllogistic reasoning, a classical testbed for deductive competence, where models exhibit \emph{content effect}: a well-documented phenomenon in human cognition whereby the perceived plausibility of a conclusion overrides the logical validity of the premises.

The conflict between semantic heuristics and logical rigour is not merely an occasional error, but a structural failure mode.  Consider the contrasting syllogistic cases:
\begin{tcolorbox}[
    colback=white,
    colframe=gray!70!black,
    boxrule=0.2pt,
    arc=0pt,
    width=\columnwidth,
]
\small
\textbf{Valid Implausible}\\
\emph{Premise 1:} All things that have fins live in the desert.\\
\emph{Premise 2:} Dolphins have fins.\\
\emph{Conclusion:} Therefore, dolphins live in the desert.\\\\
\textbf{Invalid Plausible}\\
\emph{Premise 1:} All flowers need water.\\
\emph{Premise 2:} Roses need water.\\
\emph{Conclusion:} Therefore, roses are flowers.
\label{fig:syllogisms}
\end{tcolorbox}

In \emph{Valid–Implausible} arguments, the conclusion follows deductively from the premises but conflicts with world knowledge; in \emph{Invalid–Plausible} arguments, the conclusion is factually acceptable yet unsupported by the premises. In both cases, LLMs tend to align their judgements with semantic plausibility, often misclassifying valid arguments as invalid and accepting invalid ones as valid \cite{valentino2025mitigating}. These behaviours indicate that models implicitly introduce semantic constraints into formal deduction, effectively fabricating logical flaws when conclusions contradict prior knowledge.

\begin{figure*}[t]
    \centering
    \includegraphics[width=1\linewidth]{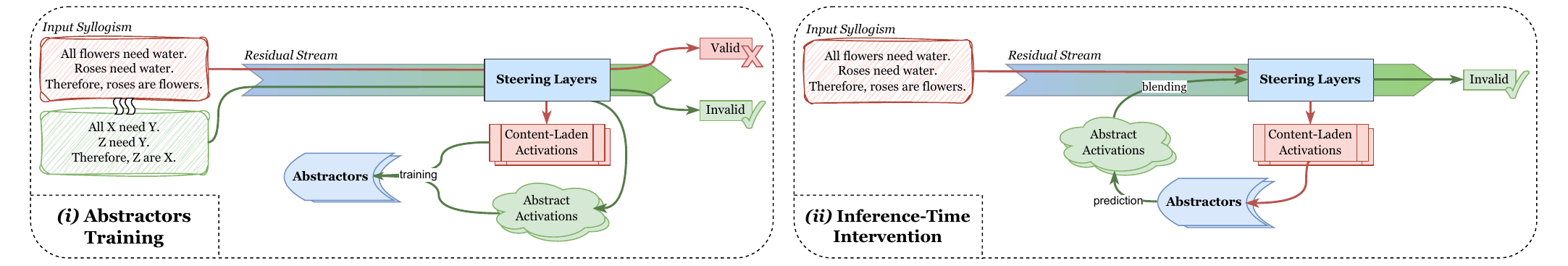}
    \caption{Overview of the abstraction steering framework. \textit{(i)} The model processes both content-laden ("All \textbf{flowers} need \textbf{water} ...") and abstract ("All \textbf{X} need \textbf{Y} ...") syllogisms. Abstractors learn to map content-laden activations into the abstract reasoning space, \textit{(ii)} integrating predicted targets via multi-layer activation at inference-time.}
    \label{fig:overview}
\end{figure*}

Existing strategies for mitigating content effect expose a trade-off between cost, modularity, and reliability. Supervised fine-tuning can improve accuracy but is computationally expensive and prone to reinforcing superficial correlations instead of inducing stable logical abstractions \cite{bertolazzi-etal-2024-systematic} while CoT-based approaches offer an affordable alternative, yet often reproduce the same semantic intuitions within intermediate reasoning steps, failing to override implausible but deductively valid conclusions \cite{wan2025srfotsyllogisticreasoningframeworkthought}. Moreover, these approaches guarantee a separation between logical structure and semantic content at the level of internal representations.

These observations suggest that the core issue lies not in the representational capacity of LLMs, but in how semantic and structural signals are routed during inference. Motivated by insights from mechanistic interpretability \cite{kim-etal-2025-reasoning}, we investigate \textbf{activation steering} as an inference-time intervention \cite{rimsky-etal-2024-steering,turner2024steeringlanguagemodelsactivation,lucchetti2025understandingcodellmsmispredicttypes}. 
By selectively intervening on the residual stream, activation steering enables the attenuation of semantic content signals while amplifying representations associated with formal structure.

We introduce \textbf{Abstractors}, lightweight Multi-Layer Perceptrons trained to map content-laden activations (last token activation of input "All \textbf{flowers} need \textbf{water}") onto an \textit{abstract reasoning manifold} ("All \textbf{X} need \textbf{Y}"), dynamically predicting the target activation for each input, unlike static steering vectors (see Figure \ref{fig:overview}). Furthermore, we evaluate the zero-shot, cross-lingual transfer of this manifold by training Abstractors exclusively on English data and evaluating them on nine other languages, both high- and low-resource ones.
In particular, to assess the generality of the proposed approach, we evaluate cross-lingual transfer by training Abstractors on English and testing them on nine additional languages, spanning both high-resource languages, such as Chinese and lower-resource languages, such as Telugu. This setting allows us to probe whether abstraction-aligned steering operates independently of linguistic surface form.

Overall, our contributions are as follows:
\begin{itemize}[noitemsep, topsep=0pt]
    \item We achieve validity-sensitive performance comparable to, and in several cases exceeding, state-of-the-art parameter-efficient methods, without modifying weights.
    \item We provide evidence that logical structure is encoded in a generalisable subspace, enabling inference-time transfer across unseen languages. 
    \item Our approach offers a modular and interpretable inference-time intervention that can be enabled or disabled on demand, preserving general language-modelling capabilities while specifically targeting belief bias.
\end{itemize}

Our results position activation-level abstraction as a scalable and language-agnostic mechanism for enhancing the robustness of formal reasoning in LLMs against semantic interference.

\section{Method}
\subsection{Overview \& Problem Formulation}

Let $x$ be a syllogism in natural language. The task is to determine its formal validity $y \in \{\texttt{valid}, \texttt{invalid}\}$. We focus on intervening on the activations of the model's residual stream. We define the activation at layer $\ell$ and token $t$ as $a_{\ell,t}(x) \in \mathbb{R}^d$, where $d$ is the model's hidden dimension.
Our approach involves a multi-layer intervention on a selected subset of target layers, $L^* \subset L$, typically located in the middle stage of the network, as the literature suggests that these layers encode higher-level semantic representations \cite{geva2021transformerfeedforwardlayerskeyvalue}. For each layer $\ell \in L^*$, we utilise the last token activation to represent the sequence:
\begin{equation}
    a_\ell(x) = a_{\ell,|x|}(x)
\end{equation}

The intervention operates at inference-time without modifying the model weights. Instead of applying a static steering vector, we propose a method to dynamically predict a target representation $\hat{a}_{\ell}(x)$ that encodes logical structure independently of semantic content.
\subsection{Abstract Reasoning Space as Target}
The objective of our steering is to steer the model's activations toward representations that encode the logical structure of the syllogism while abstracting away semantic noise.

To define these ideal representations, we construct a parallel dataset. Each content-laden syllogism $x^{\text{con}}$ is paired with an abstract version $x^{\text{abs}}$ that preserves the logical form but replaces content words with abstract symbols (e.g., "All $A$ are $B$"). The activations produced by the model when processing these abstract versions, $a_\ell(x^{\text{abs}})$, serve as our \textbf{target space} for pure formal reasoning. We hypothesise that mapping $a_\ell(x^{\text{con}})$ onto this "\textit{abstract reasoning space}" reduces reliance on semantic heuristics.
\subsection{Unified Activation Prediction via Contrastive Learning}
\label{subsec:contrastive_learning}
At inference time, the abstract counterpart $x^{\text{abs}}$ is not available. Therefore, we train a single Multi-Layer Perceptron (MLP) per layer, called \textbf{Abstractors}, to predict the target abstract activation $\hat{a}_\ell(x)$ directly from the content-laden activation $a_\ell(x^{\text{con}})$. 
\paragraph{Architecture} For each layer $\ell \in L^*$, we train a unified MLP $f_\ell$ that handles both valid and invalid syllogisms. The network employs a two-headed architecture that decouples direction and magnitude prediction:
\begin{itemize}[noitemsep]
    \item \textbf{Shared Backbone:} A deep feed-forward network $h_\ell$ extracts features from the input activation: $z = h_\ell(a_\ell(x^{\text{con}}))$
    \item \textbf{Direction Head:} Predicts the normalized unit vector $\hat{d}_\ell = \text{normalize}(g_d(z))$
    \item \textbf{Magnitude Head:} Predicts the scalar magnitude $\hat{m}_\ell = g_m(z)$
\end{itemize}
The final predicted abstract activation is:
\begin{equation}
    \hat{a}_\ell(x) = \hat{m}_\ell \cdot \hat{d}_\ell = f_\ell(a_\ell(x^{\text{con}}))
\end{equation}
\paragraph{Contrastive Training Strategy} To train the unified model to correctly handle both validity classes, we employ a contrastive learning approach with triplet construction. For each training example $x_i^{\text{con}}$, we construct a triplet $(x_i^{\text{con}}, x_i^+, x_i^-)$ where: $x_i^+$ is a \textbf{positive target}: an abstract example with the same validity as $x_i^{\text{con}}$, $x_i^-$ is a \textbf{negative counterfactual}: an abstract example with opposite validity.
\paragraph{Adaptive Matching for Triplet Construction} To ensure high-quality training data, we employ a schema-based matching process. Let $\mathcal{C}^+ = \{x^{\text{abs}} \mid \text{model predicts } x^{\text{abs}} \text{ correctly}\}$ be the set of correctly-answered abstract examples. For each $x_i^{\text{con}}$ with validity $y_i$:

\textbf{Positive Target Selection ($x_i^+$):}
\begin{enumerate}[noitemsep]
    \item \textbf{Direct Match:} If the paired $x_i^{\text{abs}} \in \mathcal{C}^+$ and has validity $y_i$, use it directly
    \item \textbf{Schema-Based Fallback:} Otherwise, select the nearest neighbour (via cosine similarity of activations) from $\mathcal{C}^+$ with the same logical schema and validity $y_i$
    \item \textbf{Validity-Based Fallback:} If no schema match exists, select the nearest neighbour from $\mathcal{C}^+$ with validity $y_i$
\end{enumerate}

\textbf{Negative Counterfactual Selection ($x_i^-$):}
Select the nearest neighbour from $\mathcal{C}^+$ with opposite validity $\neg y_i$, without schema constraints.
\paragraph{Combined Loss Function} The training objective combines three losses:
\begin{equation}
    \mathcal{L}_{\text{total}} = \mathcal{L}_{\text{attract}} + \lambda_{\text{repel}} \cdot \mathcal{L}_{\text{repel}} + \lambda_{\text{mag}} \cdot \mathcal{L}_{\text{mag}}
\end{equation}

where:
\begin{itemize}
    \item \textbf{Attraction Loss} aligns the predicted direction with the positive target:
    \begin{equation}
        \mathcal{L}_{\text{attract}} = \mathbb{E}_{i}\left[1 - \cos(\hat{d}_\ell(x_i^{\text{con}}), d_\ell(x_i^+))\right]
    \end{equation}
    where $d_\ell(x_i^+) = \text{normalize}(a_\ell(x_i^+))$
    
    \item \textbf{Repulsion Loss} pushes the prediction away from the negative counterfactual:
    \begin{multline}
        \mathcal{L}_{\text{repel}} = \\
        \mathbb{E}_{i}\left[\text{ReLU}\left(\cos(\hat{d}_\ell(x_i^{\text{con}}), d_\ell(x_i^-)) - \mu\right)\right]
    \end{multline}
    where $\mu$ is a margin threshold (typically 0), and $d_\ell(x_i^-) = \text{normalize}(a_\ell(x_i^-))$
    
    \item \textbf{Magnitude Loss} matches the predicted magnitude to the positive target:
    \begin{equation}
        \mathcal{L}_{\text{mag}} = \mathbb{E}_{i}\left[(\hat{m}_\ell(x_i^{\text{con}}) - \|a_\ell(x_i^+)\|_2)^2\right]
    \end{equation}
\end{itemize}

This contrastive formulation enables the unified model to learn a representation space where valid and invalid syllogisms are mapped to distinct regions, without requiring explicit class labels at inference time.
\subsection{Inference-Time Steering}
Since the target abstraction $\hat{a}_\ell(x)$ depends on the final token's activation, we employ a \textbf{two-pass inference strategy}:
\begin{enumerate}[noitemsep]
    \item \textbf{Pre-computation Pass:} We perform a standard forward pass on input $x$ to extract the unsteered activation of the final token, $a_{\ell,|x|}(x)$, and compute the target $\hat{a}_\ell(x)$ using the trained Abstractor.
    \item \textbf{Steered Pass:} We re-process $x$. During this second pass, the pre-computed target $\hat{a}_\ell(x)$ is blended into the stream as described below.
\end{enumerate}
\paragraph{Steering Layers} We steer multiple contiguous layers rather than a single layer to ensure robust propagation of the abstract representation throughout the forward pass. Target layers $L^*$ are selected by identifying regions where positive and negative abstract targets exhibit maximal separation (lowest cosine similarity), typically occurring in the middle layers of the network where abstract concepts have formed but remain distinct. Appendix~\ref{app:posneg_cossim} visualizes this layer-wise separation analysis across all models, with the selected steering layers corresponding to regions of minimal positive-negative similarity (full details in Appendix \ref{app:layers}).
\paragraph{Positional Blending} The intervention intensity is modulated by a positional coefficient $\alpha_t$ that increases linearly along the sequence:

\begin{equation}
    \alpha_t = \begin{cases}
        0 & \text{if } t < t_{\text{start}} \\
        \alpha \cdot \frac{t - t_{\text{start}}}{T - t_{\text{start}}} & \text{if } t \geq t_{\text{start}}
    \end{cases}
\end{equation}

where $\alpha \in [0,1]$ is the maximum steering strength, $T$ is the sequence length, and $t_{\text{start}}$ is the token position where the instruction ends and the actual syllogism content begins.
\paragraph{Activation Blending} The steered activation for token $t$ at layer $\ell$ is computed as:
\begin{equation}
    a^{\text{steer}}_{\ell,t}(x) = (1 - \alpha_t) \cdot a_{\ell,t}(x) + \alpha_t \cdot \hat{a}_{\ell}(x)
\end{equation}
Crucially, during the Steered Pass, the single target vector $\hat{a}_\ell(x)$ (computed from the Pre-computation Pass) is broadcast to all content tokens $t \geq t_{\text{start}}$ to align the entire reasoning sequence with the abstract manifold.
\section{Experimental Setup}
\subsection{Models}\label{sec:exp_set_models} We evaluate the steering pipeline on three families of open-weights LLMs. To demonstrate how our approach scales with model capabilities, we select different models: \texttt{Qwen-2.5-7B} \cite{qwen2025qwen25technicalreport} and \texttt{3-14B} \cite{yang2025qwen3technicalreport}; \texttt{Gemma-2-9B} \cite{gemmateam2024gemma2} and \texttt{3-12B} \cite{gemmateam2025gemma3technicalreport}; \texttt{Mistral-7B-v0.3} \cite{jiang2023mistral7b} and \texttt{Ministral-3-14B} \cite{mistralai2025ministral314b}.

\subsection{Dataset}
\label{subsec:dataset}
We extend a syllogistic reasoning corpus from \cite{bertolazzi-etal-2024-systematic}, comprising 24 logical forms instantiated via taxonomic relations from WordNet. Each instance includes two premises, a conclusion, and is annotated with logical \textbf{validity} and conclusion \textbf{plausibility}.

The primary dataset is in English (\texttt{en}). To assess cross-lingual transfer, we evaluate on nine additional languages spanning diverse families, scripts, and resource levels: \textbf{High-Resource Languages (HRLs)} (French, Spanish, Italian, German, Russian, Chinese) and \textbf{Low-Resource Languages (LRLs)} (Bengali, Swahili, Telugu). The English dataset contains 2{,}780 examples, each paired with its corresponding abstract counterpart, while the datasets for the other languages contain 960 examples each. All datasets are balanced across validity and plausibility categories. Data are generated via automatic translation using GPT-4o with back-translation quality control (details in Appendix \ref{app:dataset_details}).
\subsection{Metrics}
Beyond simple accuracy, we introduce a suite of metrics designed to measure the quality and robustness of the model's reasoning process. These metrics quantify heuristic reliance, penalise it in a holistic score, and measure the effectiveness of steering intervention.

\paragraph{Belief Bias ($\Delta_\text{belief}$)} How much the plausibility of the conclusion influences the model's accuracy. To calculate this, we group the dataset into two categories:
\textbf{Belief-Consistent}: Inputs where formal validity aligns with real-world plausibility (i.e., Valid-Plausible and Invalid-Implausible cases).
\textbf{Belief-Conflict}: Inputs where logic and plausibility are at odds (i.e., Valid-Implausible and Invalid-Plausible cases).

The Belief Bias is the performance gap between these two groups, isolating the model's reliance on semantic heuristics.
\begin{align}
    \text{Acc}_\text{consistent} &= \text{Acc}_{x \in \{v \iff p\}} \\
    \text{Acc}_\text{conflict} &= \text{Acc}_{x \in \{v \iff \neg p\}} \\
    \Delta_\text{belief} &= | \text{Acc}_\text{consistent} - \text{Acc}_\text{conflict} |
\end{align}
\paragraph{Bias-Penalized Accuracy (BPA)}
Global accuracy alone can be misleading, as a model might achieve high performance by exploiting heuristics. We propose a holistic metric, Bias-Penalised Accuracy, that directly penalises the model for relying on semantic content. The BPA scales the model's global accuracy by its robustness to belief bias.
\begin{equation}
    \text{BPA} = \text{Acc}_\text{global} \times (1 - \Delta_\text{belief})
\end{equation}

\paragraph{Abstract Alignment ($\eta$)}
To evaluate our hypothesis that steering maps content-laden representations onto an abstract manifold, we measure how closely the steered model approaches the performance obtained on the subset of purely abstract syllogisms:
\begin{equation}
    \eta = \frac{\text{Acc}_\text{steered}}{\text{Acc}_\text{abstract}},
\end{equation}
where $\eta = 1.0$ indicates the steered model matches the abstract upper bound, $\eta < 1.0$ indicates performance below the bound, and $\eta > 1.0$ indicates the steered model exceeds the abstract upper bound.
\subsection{Evaluation}
\paragraph{Cross-Validation} To ensure reliable results, all Abstractors are trained and evaluated using a 3-fold stratified cross-validation scheme. The data are stratified by syllogism validity to maintain an equal class distribution within each fold. All reported results are the average across these three folds.

\paragraph{Cross-Lingual Transfer} Our primary hypothesis is that the Abstractors learn a generalisable representation of logic across languages. To test this, Abstractors are trained exclusively on the English set, and their performance is evaluated in a zero-shot way on the test sets for English, French, Spanish, Italian, and Chinese.

\paragraph{Steering Strength Ablation} We conduct an ablation study on the maximum steering strength hyperparameter, $\alpha$, testing values from $\{0.1, 0.2, ..., 1.0\}$. The optimal $\alpha$, for each model, is selected based on the highest BPA on the English validation set, and this value is used for all results. Detailed performance for each $\alpha$ value are shown in Appendix \ref{app:strablation}.
\paragraph{Baselines} To evaluate the effectiveness of inference-time intervention, we compare our with: Baseline (No Steering): The original model's zero-shot performance. Supervised Fine-Tuning (SFT): A parameter-efficient fine-tuning baseline. To ensure a fair comparison, we employ PiSSA (PrIncipal Singular values and Singular vectors Adaptation) \cite{meng2025pissaprincipalsingularvalues}, a state-of-the-art PEFT method that outperforms standard LoRA by initializing adapters via SVD. Implementation details are discussed in Appendix \ref{app:sft}. Chain-of-Thought (CoT): To evaluate whether the model's logical reasoning can be elicited purely through prompt engineering, we employ Chain-of-Thought \cite{wei2023chainofthoughtpromptingelicitsreasoning}. Details are discussed in Appendix \ref{app:cot}.

Due to computational constraints, baseline methods were evaluated on a subset of the cross-validation folds. Specifically, we trained SFT adapters for three models (\texttt{Mistral-7B}, \texttt{Qwen-2.5-7B}, and \texttt{Gemma-2-9B}) on Fold~0 only. For consistency, the same restriction was applied to CoT. Accordingly, SFT and CoT results report performance for these three models on Fold~0, while Steering results are averaged across all three folds for all models. A direct comparison with Steering evaluated on Fold~0 can be found in Appendix~\ref{app:steering vs sft vs cot}.
\section{Results \& Analysis}
\label{sec:results}

\begin{table*}[t]
\centering
\renewcommand{\arraystretch}{1.2}
\resizebox{\textwidth}{!}{
\begin{tabular}{l|cc|cc|cc||cc|cc|cc||cc|cc|cc}
\toprule
& \multicolumn{6}{c||}{\textbf{English}} & \multicolumn{6}{c||}{\textbf{HRLs}} & \multicolumn{6}{c}{\textbf{LRLs}} \\
\cmidrule(lr){2-7} \cmidrule(lr){8-13} \cmidrule(lr){14-19}
\multirow{2}{*}{\textbf{Model}} 
& \multicolumn{2}{c|}{\textbf{Base Model}} & \multicolumn{2}{c|}{\textbf{SFT}} & \multicolumn{2}{c||}{\textbf{CoT}} 
& \multicolumn{2}{c|}{\textbf{Base Model}} & \multicolumn{2}{c|}{\textbf{SFT}} & \multicolumn{2}{c||}{\textbf{CoT}}
& \multicolumn{2}{c|}{\textbf{Base Model}} & \multicolumn{2}{c|}{\textbf{SFT}} & \multicolumn{2}{c}{\textbf{CoT}} \\
\cmidrule(lr){2-3} \cmidrule(lr){4-5} \cmidrule(lr){6-7} \cmidrule(lr){8-9} \cmidrule(lr){10-11} \cmidrule(lr){12-13} \cmidrule(lr){14-15} \cmidrule(lr){16-17} \cmidrule(lr){18-19}
& BPA & →\textbf{Steer} & BPA & $\Delta_B$ & BPA & $\Delta_B$ 
& BPA & →\textbf{Steer} & BPA & $\Delta_B$ & BPA & $\Delta_B$
& BPA & →\textbf{Steer} & BPA & $\Delta_B$ & BPA & $\Delta_B$ \\
\midrule
\texttt{Qwen-2.5-7B} & 50.8 & \textbf{95.4} & 98.6 & \textcolor{Green}{+3.2} & 81.7 & \textcolor{Red}{-13.7} & 42.9 & \textbf{86.1} & 84.6 & \textcolor{Red}{-1.5} & 57.7 & \textcolor{Red}{-28.4} & 55.2 & \textbf{64.8} & 70.1 & \textcolor{Green}{+5.3} & 55.0 & \textcolor{Red}{-9.8} \\
\texttt{Qwen-3-14B} & 72.3 & \textbf{97.8} & 95.8 & \textcolor{Red}{-2.0} & 83.1 & \textcolor{Red}{-14.7} & 57.5 & \textbf{86.9} & -- & -- & -- & -- & 54.0 & \textbf{70.8} & -- & -- & -- & -- \\
\texttt{Gemma-2-9B} & 51.8 & \textbf{98.0} & 98.8 & \textcolor{Green}{+0.8} & 69.9 & \textcolor{Red}{-28.1} & 43.9 & \textbf{89.0} & 93.4 & \textcolor{Green}{+4.4} & 58.5 & \textcolor{Red}{-30.5} & 47.0 & \textbf{73.5} & 77.0 & \textcolor{Green}{+3.5} & 53.6 & \textcolor{Red}{-19.9} \\
\texttt{Gemma-3-12B} & 79.4 & \textbf{97.0} & 97.7 & \textcolor{Green}{+0.7} & 86.8 & \textcolor{Red}{-10.2} & 63.5 & \textbf{90.1} & -- & -- & -- & -- & 58.0 & \textbf{79.0} & -- & -- & -- & -- \\
\texttt{Mistral-7B} & 30.2 & \textbf{96.1} & 88.1 & \textcolor{Red}{-8.0} & 34.3 & \textcolor{Red}{-61.8} & 30.0 & \textbf{75.4} & 78.2 & \textcolor{Green}{+2.8} & 34.2 & \textcolor{Red}{-41.2} & 47.6 & \textbf{53.0} & 53.7 & \textcolor{Green}{+0.7} & 50.0 & \textcolor{Red}{-3.0} \\
\texttt{Ministral-14B} & 63.9 & \textbf{97.6} & 96.3 & \textcolor{Red}{-1.3} & 76.2 & \textcolor{Red}{-21.4} & 49.2 & \textbf{86.1} & -- & -- & -- & -- & 50.4 & \textbf{69.2} & -- & -- & -- & -- \\
\midrule
\textbf{Average} & 58.1 & \textbf{97.0} & 95.9 & \textcolor{Red}{-1.1} & 72.0 & \textcolor{Red}{-25.0} & 47.8 & \textbf{85.6} & 85.4 & \textcolor{Red}{-0.2} & 50.1 & \textcolor{Red}{-35.5} & 52.0 & \textbf{68.4} & 66.9 & \textcolor{Red}{-1.5} & 52.9 & \textcolor{Red}{-15.5} \\
\bottomrule
\end{tabular}%
}
\caption{Comparison of Bias-Penalized Accuracy (BPA) across all methods and language groups. \textbf{Steering} (bold) shows average performance across 3 folds for all models. $\Delta_B$ shows the difference relative to Steering (\textcolor{Green}{green} = improvement, \textcolor{Red}{red} = degradation).
}
\label{tab:main_results}
\end{table*}

\subsection{Main Results: Steering Enhances Accuracy and Reduces Bias}
\begin{figure}[h]
    \centering
    \includegraphics[width=1.0\columnwidth]{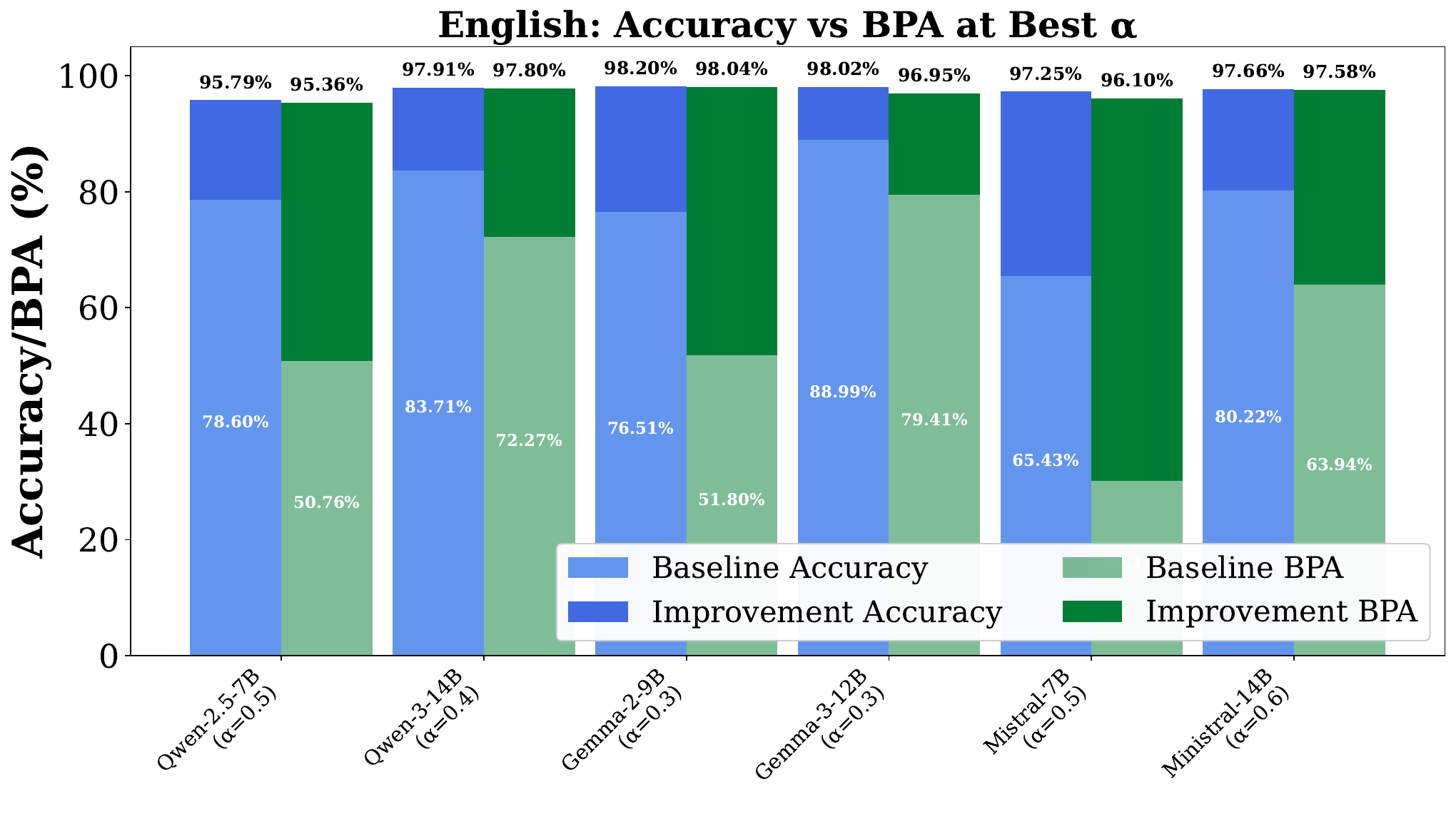}
    \caption{Performance on the English test set. Global Accuracy (blue) and Bias-Penalised Accuracy (green); darker shades indicate gains from activation steering. 
    }
    \label{fig:bpa_en}
\end{figure}

We first evaluate the effectiveness of the Abstractors on the English test set. Table \ref{tab:main_results} shows the Bias-Penalized Accuracy (BPA) increase for all models, with Figure \ref{fig:bpa_en} providing a visual breakdown.

The results demonstrate consistent improvements across all model families. Mapping content-laden activations towards an abstract reasoning space yields BPA gains ranging from $+16.54$ points (\texttt{Gemma-3-12B}) to $+42.99$ points (\texttt{Qwen-2.5-7B}). The weaker baseline models exhibit the most dramatic improvements, with \texttt{Mistral-7B} achieving a $+65.94$ pp increase in BPA, transforming a highly biased baseline ($30.16$) into a competent reasoner ($96.10$).

The improvements in BPA consistently exceed those in raw Global Accuracy. For instance, whilst \texttt{Qwen-2.5-7B} sees an accuracy gain of $+15.21$ pp, its BPA jumps by $+42.99$ pp. This disparity reveals that steering does not merely improve the model's thinking strategy; rather, it fundamentally alters the reasoning process by suppressing reliance on semantic content: precisely the type of heuristic dependence that BPA is designed to penalise. This pattern holds across all models: the reduction in Content Effect (reflected in the larger BPA gains) indicates that the Abstractors successfully dise3ntangle logical form from semantic content, forcing the model to reason about structure. 

\subsection{Cross-Lingual Generalisation}
\label{subsec:crosslingual}
\begin{figure}[h]
    \centering
    \includegraphics[width=1.0\columnwidth]{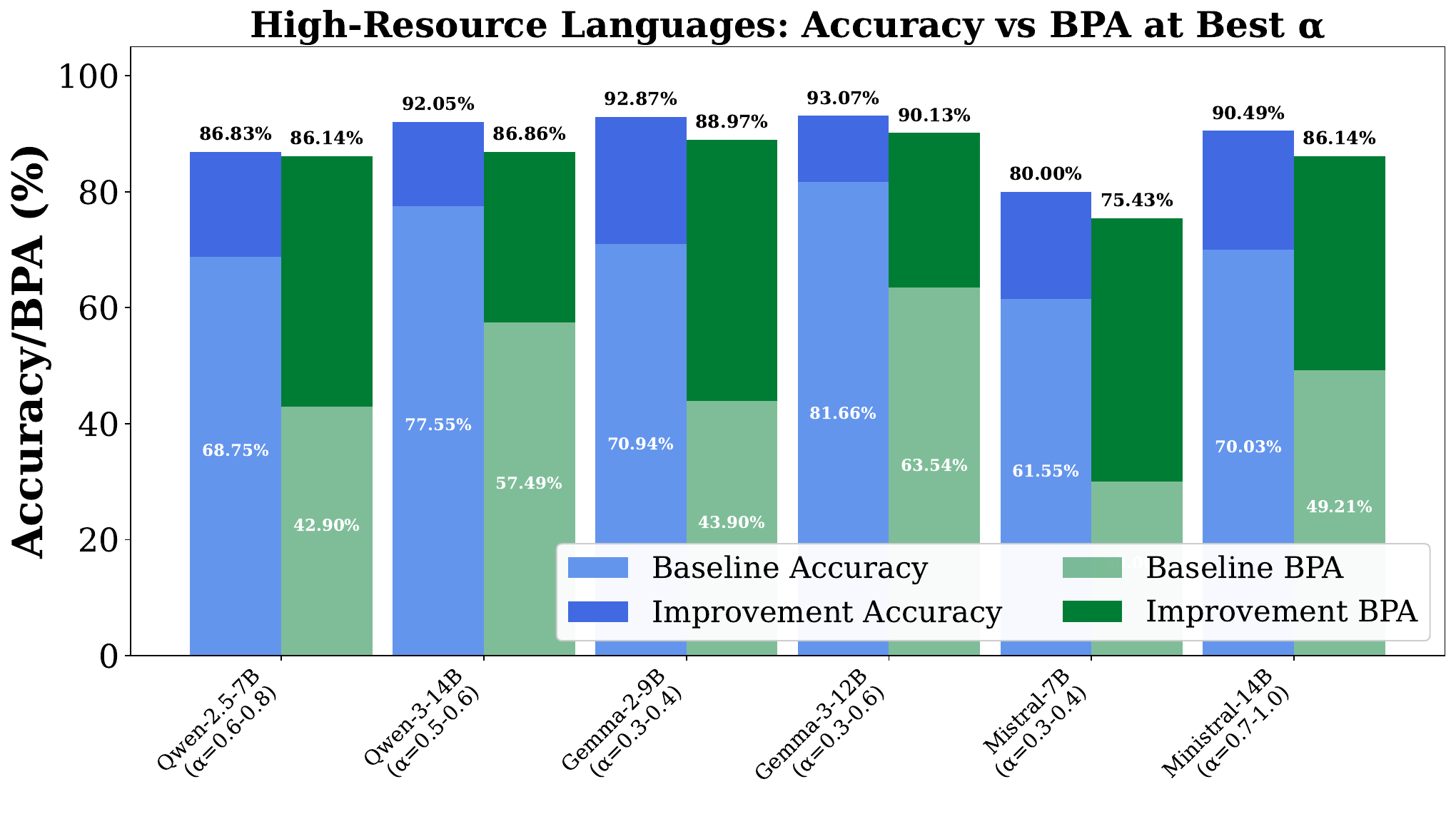}
    \caption{Performance comparison on HRLs (French, Spanish, Italian, German, Russian, Chinese).
    }
    \label{fig:bpa_hrl}
\end{figure}

We now test a central hypothesis of this work: that the Abstractors capture a generalisable "\textit{logic manifold}" that transcends language. 
To provide a structured analysis of transfer limits, we partition evaluation languages into HRLs and LRLs (as mentioned in \S \ref{subsec:dataset}). This stratification allows us to disentangle the effects of the steering mechanism from the quality of the base model's pre-trained representations.
\paragraph{HRLs: Near-Perfect Transfer}
Table \ref{tab:main_results} and Figure \ref{fig:bpa_hrl} reveal robust transfer across all model families. Steering achieves percentage increases in BPA that often exceed those observed in English.

The consistency of these results is significant. For instance, \texttt{Gemma-3-12B} achieves an average BPA of $90.13$ across HRLs, only slightly below its English performance of $96.95$. This transfer extends even to languages with fundamentally different scripts and tokenisation schemes such as Chinese, which relies on logographic characters and entirely different subword structure. This suggests the Abstractors target a deeper semantic layer encoding logical structure in a manner coherent across typologically diverse languages, rather than exploiting superficial token-level patterns.

\begin{figure}[h]
    \centering
    \includegraphics[width=1.0\columnwidth]{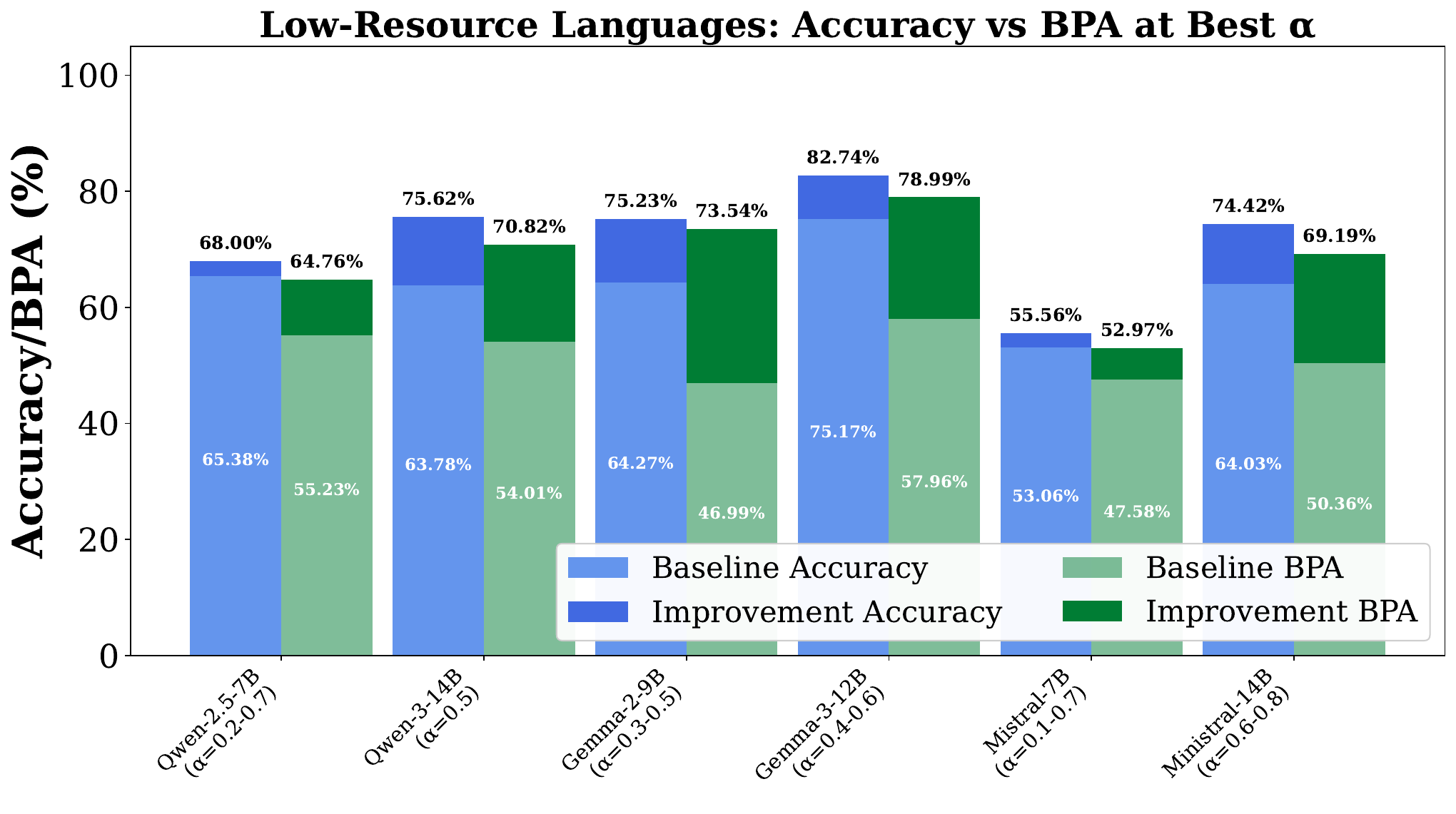}
    \caption{Performance comparison on LRLs (Bengali, Swahili, Telugu). 
    }
    \label{fig:bpa_lrl}
\end{figure}

\paragraph{LRLs: Meaningful but Limited Gains}
Whilst steering consistently improves performance, Table \ref{tab:main_results} and Figure \ref{fig:bpa_lrl} show that the absolute gains are noticeably smaller.

BPA improvements in LRLs range from $+5.39$ pp (\texttt{Mistral-7B}) to $+26.55$ pp (\texttt{Gemma-2-9B}). Whilst substantial in relative terms--often representing $+30\%$ to $+50\%$ increases--they fall short of the near-perfect alignment observed in HRLs. This gap reflects a \emph{geometry gap} in base model representations: LRLs receive less exposure during pre-training, leading to noisier, less structured activations. Since steering amplifies and redirects existing latent patterns rather than inducing new knowledge, it cannot produce robust logical reasoning where the underlying representation quality is poor.
Still, that steering yields \textit{any} improvement in LRLs is non-trivial. Abstractors trained without encountering any tokens of Swahili, Bengali, or Telugu still identify and enhance reasoning structures in these languages, suggesting that models maintain some cross-lingual coherence in their logical-reasoning manifolds.

\noindent \textbf{Summary:}
The cross-lingual results reveal a clear hierarchy:
\emph{English} shows near-perfect steering;
\emph{HRLs} exhibit strong zero-shot transfer, remaining within $5$–$10$ pp of English;
\emph{LRLs} display meaningful but constrained improvements, reflecting weaker base representations.
This pattern provides compelling evidence that Abstractors learn a generalisable logical transformation. 

\subsection{Abstract Alignment Analysis}
Beyond performance metrics, we validate the geometric mechanism of our intervention using the Abstract Alignment ($\eta$) metric, which measures how closely steered representations approach the theoretical upper bound of purely symbolic reasoning. Detailed results in Appendix~\ref{app:abstract_results}.

\begin{table}[h]
\centering
\begin{tabular}{lccc}
\toprule
\textbf{Model} & \textbf{English} & \textbf{HRLs} & \textbf{LRLs} \\
\midrule
\texttt{Qwen-2.5-7B} & 1.15 & 1.05 & 0.83 \\
\texttt{Qwen-3-14B} & 1.09 & 1.03 & 0.85 \\
\texttt{Gemma-2-9B} & 1.17 & 1.11 & 0.90 \\
\texttt{Gemma-3-12B} & 1.04 & 0.99 & 0.89 \\
\texttt{Mistral-7B} & 1.14 & 0.96 & 0.65 \\
\texttt{Ministral-14B} & 1.15 & 1.07 & 0.88 \\
\bottomrule
\end{tabular}
\caption{Abstract Alignment ($\eta$) across language groups. Values represent the ratio of steered accuracy to abstract upper bound accuracy at the best BPA $\alpha$.}
\label{tab:abstract_alignment}
\end{table}

\noindent \textbf{English Alignment.}
English models exhibit strong alignment, with $\eta$ values consistently exceeding $1.0$ (Table \ref{tab:abstract_alignment}). This confirms that steering vectors bridge the gap between content-laden and abstract activation manifolds, improving performance by reducing noise in the Abstractor network.

\noindent \textbf{High-Resource Languages (HRLs).}
The inference-time nature of our approach is most evident in HRLs. Despite Abstractors being trained solely on English, $\eta$ scores remain clustered around $1.0$ across all HRLs, demonstrating that they share an essentially isomorphic reasoning geometry with English in the model's latent space.

\noindent \textbf{Low-Resource Languages (LRLs).}
In LRLs, alignment remains high but shows slight degradation. The variance in $\eta$ scores suggests the target abstract manifold is intrinsically noisier due to limited pre-training exposure, consistent with our earlier findings on the \textit{geometry gap} (\S \ref{subsec:crosslingual}).

\subsection{Comparative Analysis}
We situate Activation Steering by comparing it with Chain-of-Thought reasoning and weight-based SFT, detailed results in Appendix \ref{app:steering vs sft vs cot}.
Contrary to trends in general reasoning tasks, CoT yields the lowest performance among methods and significantly underperforms steering in every scenario, with relative BPA differences ranging from $-11.55\%$ to $-63.86\%$. 
SFT via PiSSA represents the upper bound of standard interventions. SFT generally performs on par with, or slightly better than, steering in \texttt{Qwen} and \texttt{Gemma}. Relative BPA differences with respect to steering only range from $-7.05\%$ to $+7.38\%$. Yet, this comes at the cost of modularity: SFT alters model weights, risking catastrophic forgetting or overfitting to the syllogistic format. Notably, for \texttt{Mistral-7B}, steering outperforms SFT, suggesting that, for some architectures, reasoning circuits are better accessed via inference-time geometry than weight modification.

Activation steering settles a unique position, balancing accuracy, control, and modularity. Unlike SFT, which operates as black-box optimisation and may reinforce superficial shortcuts, steering provides a mechanistically grounded intervention that targets the internal representation of logical validity. This results in substantial performance gains over both CoT and SFT (Appendix \ref{app:steering vs sft vs cot}, Table~\ref{tab:steering vs sft vs cot}), indicating that Content Effect reflects a failure of control. Moreover, steering requires no weight updates: Abstractors are lightweight and can be toggled at inference time, preserving the model's general behaviour.

\subsection{Fluency Sanity Check}
\label{subsec:ppl}
A risk of training Abstractors in English is that they may implicitly bias the model toward English-centric representations, which could manifest as increased perplexity (PPL) in other languages. We measure the relative increase in perplexity across all languages on mC4 \cite{xue-etal-2021-mt5} as a function of steering strength $\alpha$.
At the optimal $\alpha$ identified in \S \ref{sec:results} ($\alpha \approx 0.4 - 0.6$), the PPL increase remains moderate ($\le 5\%$). More significant degradation ($>10\%$) is observed only at aggressive steering levels ($\alpha \ge 0.8$), where the abstract representations begin to surpass the syntactic features necessary for coherent generation. PPL increases uniformly across languages, suggesting no English-specific bias and confirming that the method operates safely, correcting reasoning without altering models' generative capabilities (details in Appendix \ref{app:ppl}).

\subsection{OOD Analysis}
\label{subsec:ood}
\begin{figure}[h]
    \centering
    \includegraphics[width=0.9\columnwidth]{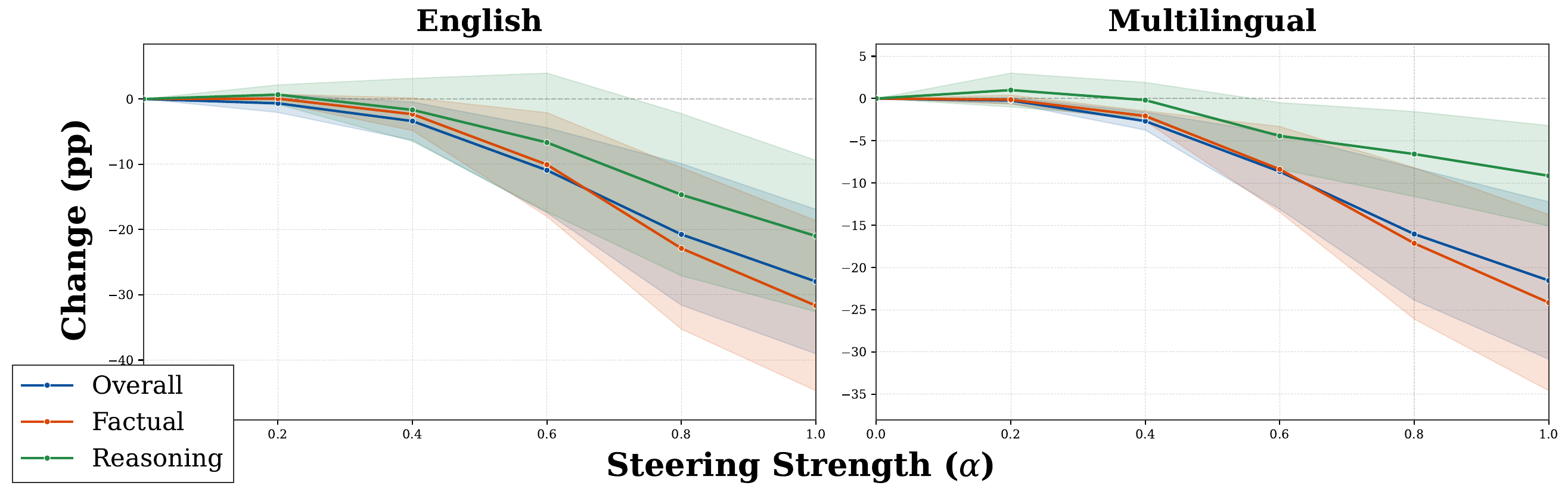}
    \caption{Performance impact on MMLU. Factual tasks (orange) and formal logic subtasks (green).
    }
    \label{fig:mmlu}
\end{figure}

We extend our evaluation to the Multilingual Massive Multitask Language Understanding (MMMLU) to assess how the abstraction mechanism impacts model capabilities. We report results in terms of accuracy and separately for two task categories: \textbf{Factual}, which includes subjects such as \textit{high school geography}, \textit{college biology}, and \textit{high school world history}; and \textbf{Reasoning}, which includes \textit{abstract algebra} and \textit{formal logic}. Since our method is designed to suppress semantic content in favour of logical form, we hypothesise a "feature suppression" effect on factual tasks.
Figure~\ref{fig:mmlu} (details Appendix~\ref{app:ood}) shows the average deltas across models and languages as a function of $\alpha$. It emerges that the Reasoning subset exhibits lower degradation than in Factual subjects, as expected given the mechanistic nature of the intervention: the specificity of the steering pipeline targets reasoning-related representations while inhibiting factual memory.
The nature of our approach allows steering to be dynamically disabled, toggled per task, or tuned to balance reasoning enhancement with general knowledge preservation.

\section{Related Work}
\label{sec:rel_works}
\noindent \paragraph{Syllogistic reasoning.}
Syllogisms provide a controlled setting for probing whether LLMs can track logical entailment independently of lexical and world-knowledge cues \cite{Evans1983,NEWSTEAD2003193}.
Although this is a formally clear task, LLMs often fail to disentangle content from form, treating semantic plausibility as evidence of validity and exhibiting biases grounded in semantic content \cite{bertolazzi-etal-2024-systematic}, rejecting valid conclusions when these disagree with background knowledge. \citet{ozeki-etal-2024-exploring,eisape-etal-2024-systematic} suggest that scaffolding paradigms yield only marginal improvements, whereas SFT reduces reliance on such heuristics but does not eliminate them \cite{ranaldi-freitas-2024-aligning}.

\noindent \textbf{Activation Steering.}
A growing body of work investigates inference-time control by intervening directly on a model’s internal activations.
These methods identify directions or subspaces associated with specific behaviours and modulate them to influence downstream predictions \citep{stoehr-etal-2024-activation,soo2025interpretablesteeringlargelanguage}. Activation steering has been applied to reduce hallucinations, mitigate harmful outputs, and improve factual reliability, offering a lightweight, modular, and reversible alternative to parameter updates.

\noindent \textbf{Abstraction for content-invariant reasoning.}
A complementary line of work improves robustness by encouraging reasoning at higher levels of abstraction.
CoT explanations can inherit semantic shortcuts that compromise faithfulness \cite{lyu-etal-2023-faithful}, motivating approaches that promote intermediate representations separating variables and relations from surface content \cite{ranaldi-etal-2025-improving,ranaldi-pucci-2025-multilingual}. These methods rely on supervision and do not intervene on internal activations during inference.

\noindent \textbf{Our contribution.}
We unify abstraction and activation steering by treating abstraction as an explicit geometric target for inference-time control. We define an abstract reasoning space from paired content-based and abstract syllogisms and learn lightweight \emph{Abstractors} that map content-conditioned activations to representations aligned with this space. These predicted targets are integrated through multi-layer interventions during inference, without modifying model weights. We evaluate the resulting abstraction-aligned steering in a cross-lingual setting, assessing whether the induced transformation captures structural regularities beyond language-specific surface cues.

\newpage

\section{Conclusion}
We showed that abstraction-guided activation steering can reduce content effect (CE) in syllogistic reasoning by separating structural inference from semantic content at the representation level. By dynamically mapping internal states to a purely logical manifold, we observe great improvements (+65 points) in bias-penalised accuracy, neutralising the CE with zero-shot transfer to languages as diverse as Chinese and Bengali. Our results highlight activation-level abstraction as a scalable mechanism for improving the robustness of formal reasoning in LLMs.

\section*{Limitations}
We focused on syllogistic reasoning as a controlled testbed for analysing content effects. While this setting allows for fine-grained control over logical form and semantic plausibility, it limits the generalisability of the findings to more complex reasoning scenarios, such as propositional or first-order logic, multi-hop inference, or open-ended reasoning tasks.
The proposed framework relies on paired content-laden and abstract examples to define a target reasoning manifold. This design assumes that suitable abstract counterparts can be constructed reliably, an assumption that may not hold in domains where abstraction is less well-defined or inherently task-specific. Our evaluation is restricted to open-weight models, as the method requires direct access to internal activations. Thus, the approach is not directly applicable to closed-source models, in which activation-level interventions are infeasible. While this does not affect the conceptual validity of abstraction-guided steering, it constrains its practical applicability in settings dominated by proprietary systems. 
\nocite{*}
\bibliography{custom}
\clearpage
\appendix

\section{Implementation Details}
\label{app:implementation}
\paragraph{Software and Hardware} All experiments were conducted using the Hugging Face Transformers library. Both training and inference were performed using half-precision (fp16) arithmetic on an NVIDIA RTX A6000 GPU.
\paragraph{Models}
\label{app:model_info}
In our experimental setting, as introduced in \S \ref{sec:exp_set_models}, we propose different models (detailed in Table \ref{tab:versions_models}).
We choose the generation temperature for (mostly) deterministic outputs. The other parameters are left unchanged as recommended by the official resources. 
\begin{table}[h]
\small
\centering 
\begin{tabular}{l|p{3.5cm}}
\textbf{Model} & \textbf{Version}  \\ 
\hline
Qwen-2.5-7B   &  Qwen/Qwen2.5-7B-Instruct \\ 
Qwen-3-14B   & Qwen/Qwen3-14B \\
Gemma-2-9B   & google/gemma-2-9b-it \\
Gemma-3-12B   & google/gemma-3-12b-it \\
Mistral-7B   & mistralai/Mistral-7B-Instruct-v0.3 \\
Ministral-14B   & mistralai/Ministral-3-14B-Instruct-2512 \\\hline
\hline
\end{tabular}
\caption{List the versions of the models proposed in this work, which can be found on huggingface.co. We used all the default configurations proposed in the repositories for each model.}
\label{tab:versions_models}
\label{app:model_versions}
\vspace{1cm}
\end{table}
\subsection{Steering Layer Selection}
\label{app:layers}
Optimal steering layers are identified empirically by analysing the geometric separation between positive and negative target abstract representations. For each content-laden syllogism, we identify its positive target (an abstract syllogism with the same validity) and negative counterfactual (an abstract syllogism with opposite validity) using the matching procedure described in \S \ref{subsec:contrastive_learning}. 

For each layer $\ell$, we compute the average cosine similarity between these positive and negative target pairs:
\begin{equation}
    s_\ell = \frac{1}{N} \sum_{i=1}^{N} \cos(a_\ell(y_i^+), a_\ell(y_i^-))
\end{equation}
where $N$ is the number of content-laden examples, $y_i^+$ is the positive abstract target for example $i$, and $y_i^-$ is the corresponding negative counterfactual.

Layers with the lowest similarity $s_\ell$ exhibit the clearest separation between validity classes in the abstract reasoning space, making them optimal targets for steering. These layers are typically found in the middle layers (second or third quarter) of the network, where abstract concepts have formed but have not yet collapsed into task-specific outputs.
\begin{table}[h]
\centering
\small
\begin{tabular}{lcc}
\toprule
\textbf{Model}  & \textbf{Selected Layers} \\
\midrule
\texttt{Qwen-2.5-7B} & [18, 19, 20, 21, 22] \\
\texttt{Qwen-3-14B} & [24, 25, 26, 27, 28, 29] \\
\texttt{Gemma-2-9B} & [21, 22, 23, 24, 25] \\
\texttt{Gemma-3-12B} & [24, 25, 26, 27, 28, 29] \\
\texttt{Mistral-7B} & [13, 14, 15, 16, 17] \\
\texttt{Ministral-3-14B} & [20, 21, 22, 23, 24, 25] \\
\bottomrule
\end{tabular}
\caption{Optimal steering layers selected via similarity analysis (0-indexed).}
\label{tab:steering_layers_appendix}
\end{table}

\begin{figure*}[htbp]
\section{Positive-Negative Cosine Similarity Analysis}
\label{app:posneg_cossim}
    \centering
    
    \begin{subfigure}[b]{0.45\textwidth}
        \centering
        \includegraphics[width=\textwidth]{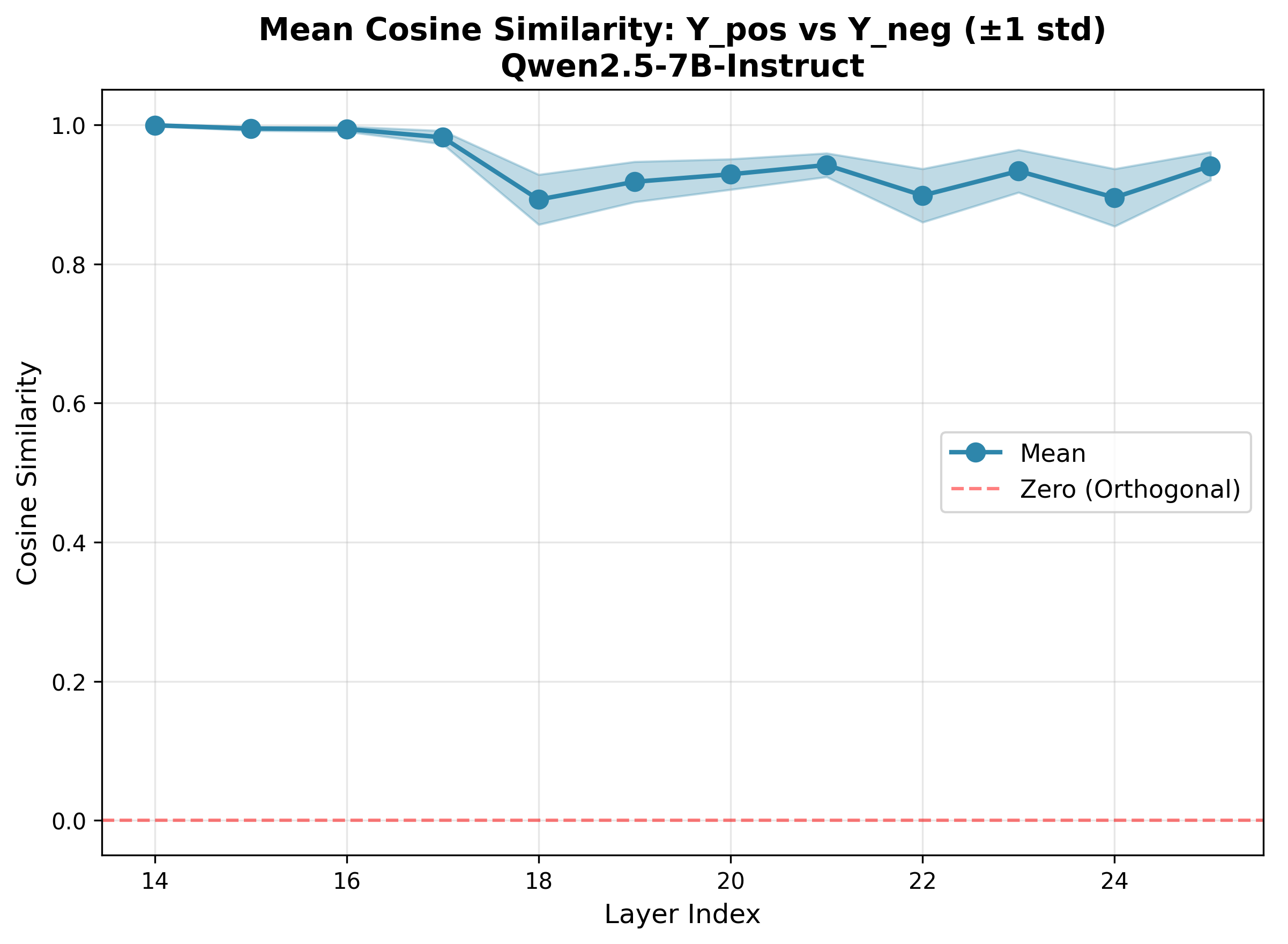}
        \caption{Qwen2.5-7B}
        \label{fig:sub1}
    \end{subfigure}
    \hfill
    \begin{subfigure}[b]{0.45\textwidth}
        \centering
        \includegraphics[width=\textwidth]{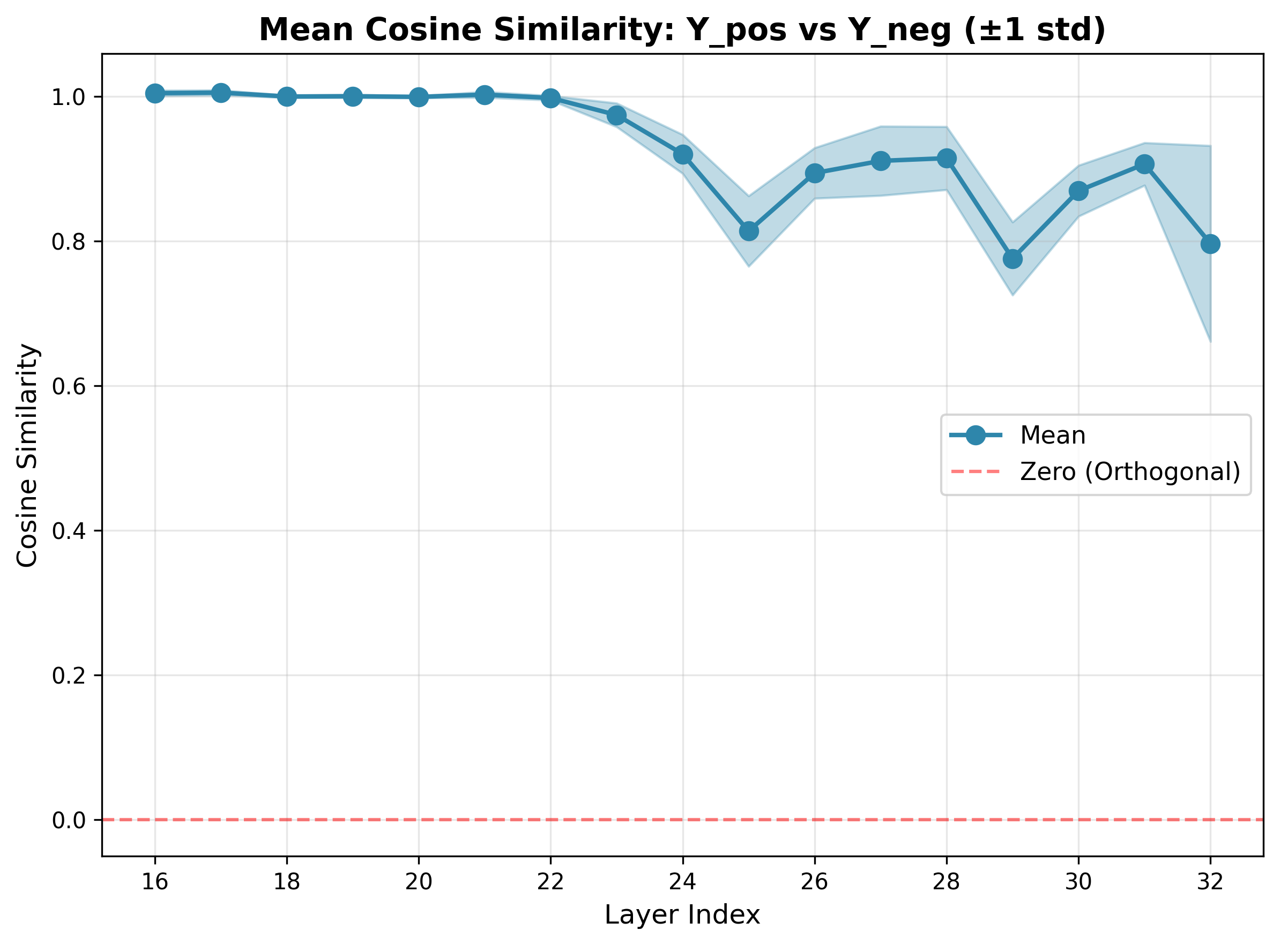}
        \caption{Qwen3-14B}
        \label{fig:sub2}
    \end{subfigure}
    
    \vspace{0.5cm}
    
    \begin{subfigure}[b]{0.45\textwidth}
        \centering
        \includegraphics[width=\textwidth]{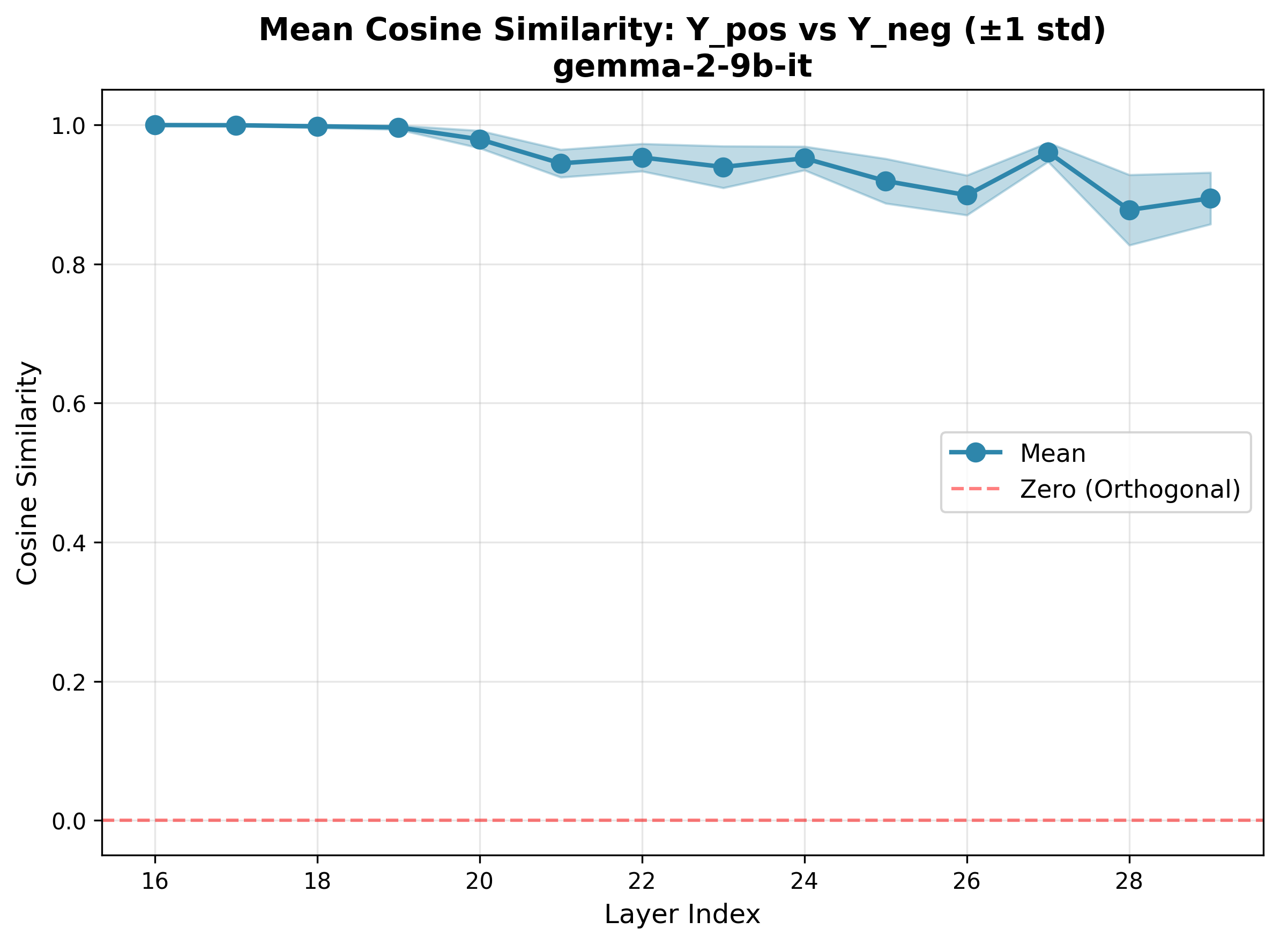}
        \caption{Gemma-2-9B}
        \label{fig:sub3}
    \end{subfigure}
    \hfill
    \begin{subfigure}[b]{0.45\textwidth}
        \centering
        \includegraphics[width=\textwidth]{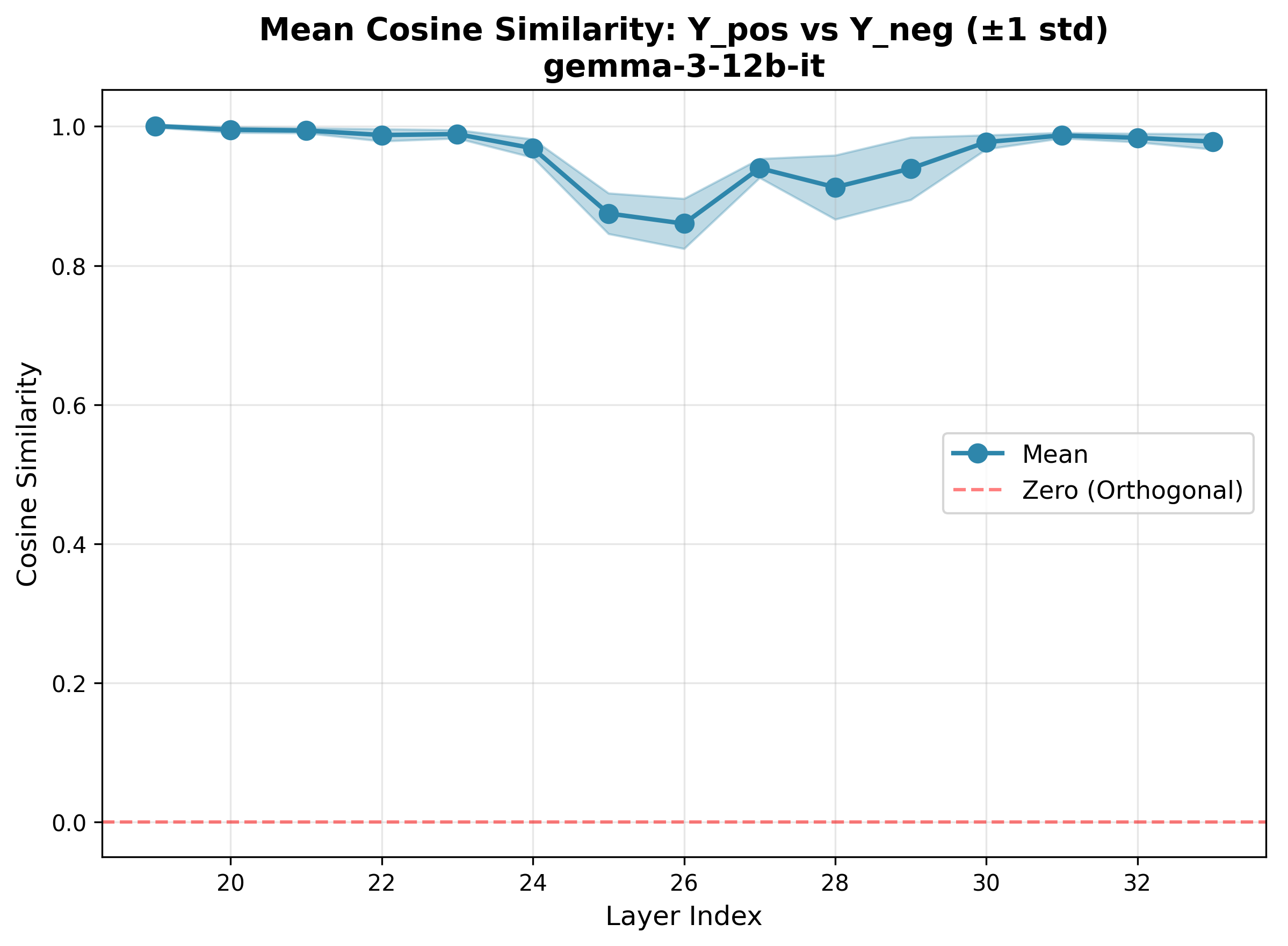}
        \caption{Gemma-3-12B}
        \label{fig:sub4}
    \end{subfigure}
    
    \vspace{0.5cm}
    
    \begin{subfigure}[b]{0.45\textwidth}
        \centering
        \includegraphics[width=\textwidth]{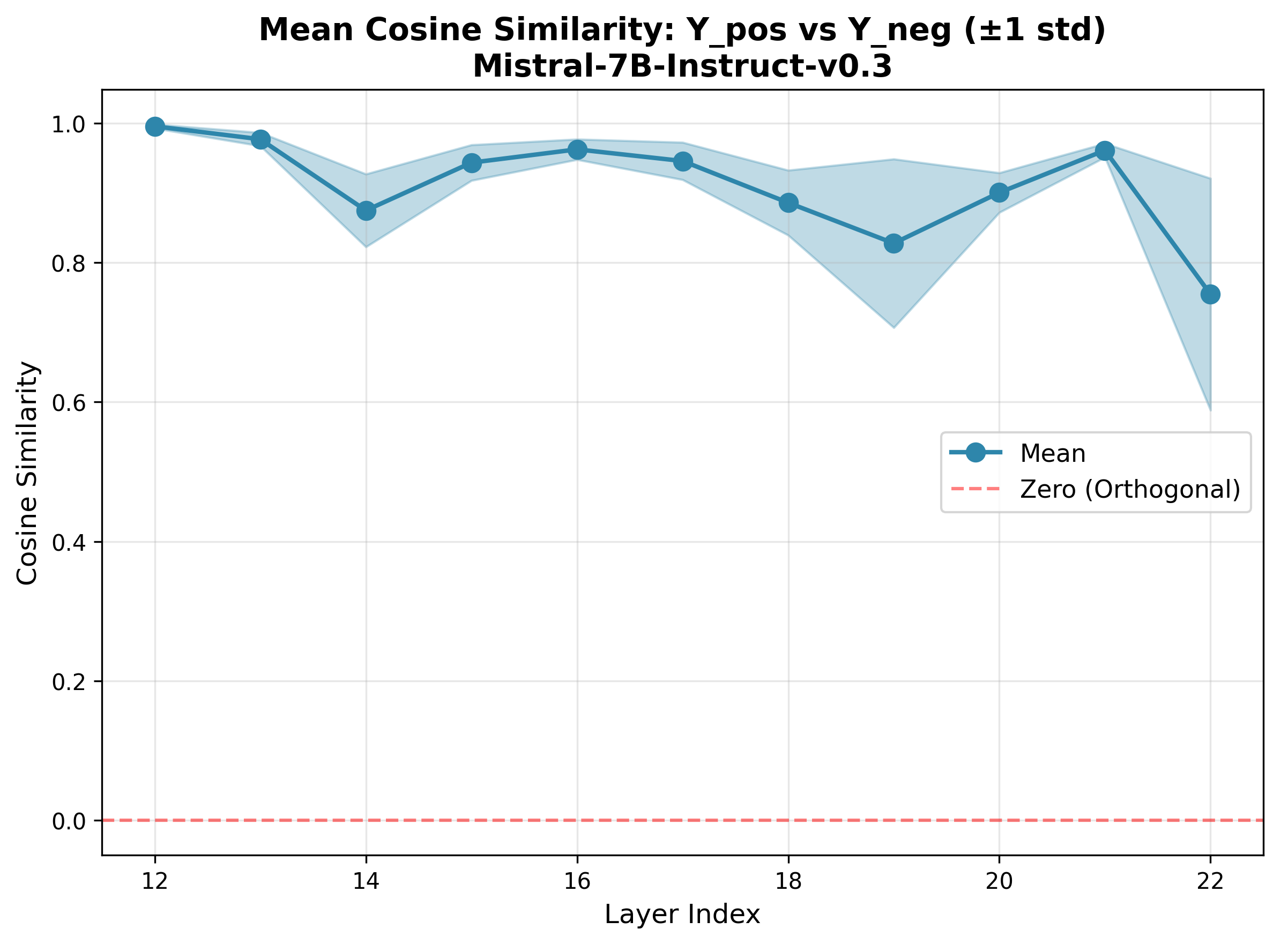}
        \caption{Mistral-7B}
        \label{fig:sub5}
    \end{subfigure}
    \hfill
    \begin{subfigure}[b]{0.45\textwidth}
        \centering
        \includegraphics[width=\textwidth]{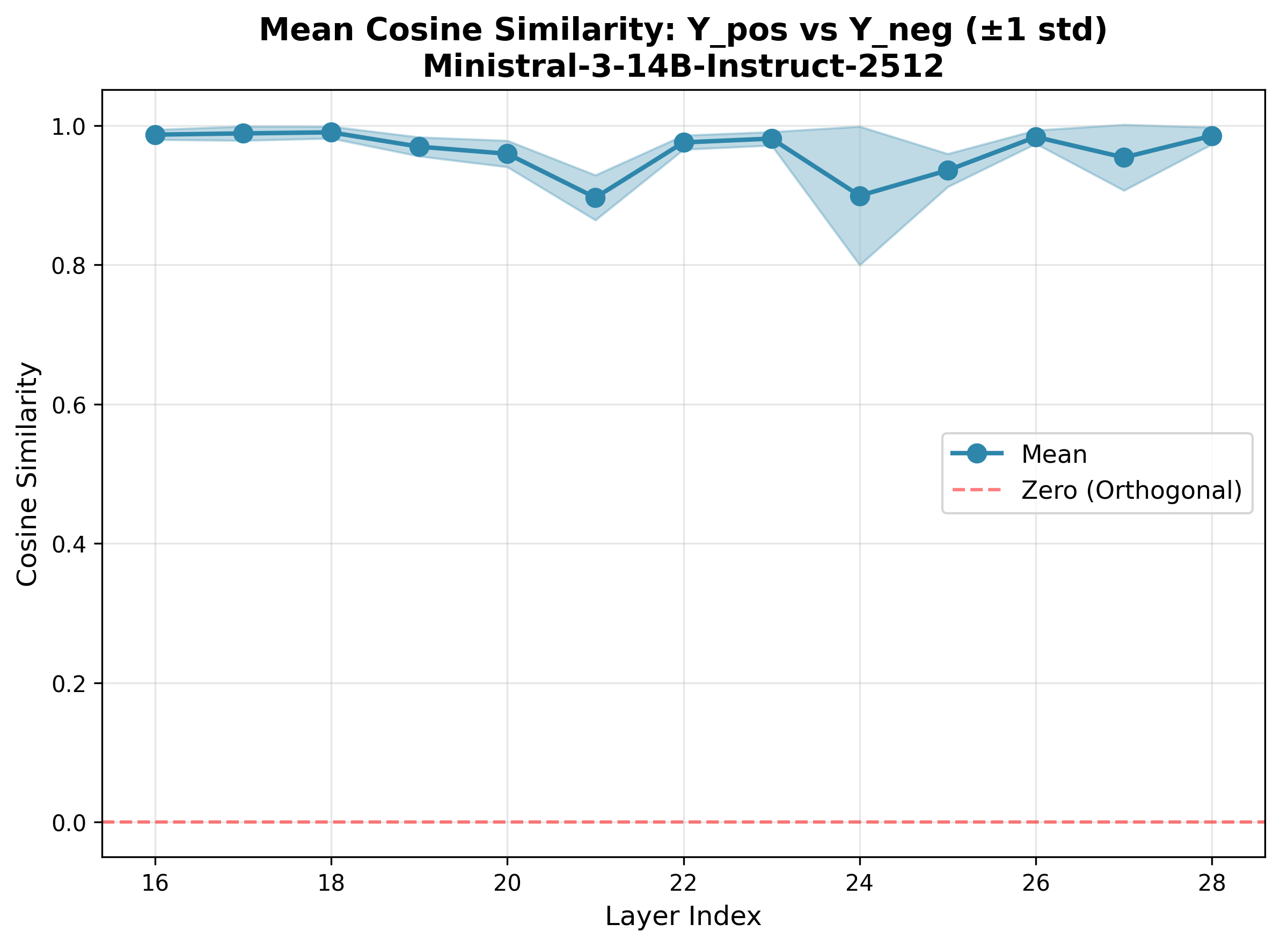}
        \caption{Ministral-3-14B}
        \label{fig:sub6}
    \end{subfigure}
    \caption{Layer-wise cosine similarity between positive and negative training target pairs across all models.}
    \label{fig:posneg_cossim}
\end{figure*}

Appendix \ref{app:posneg_cossim}, Figure \ref{fig:posneg_cossim} visualizes this layer-wise analysis across all models.
\subsection{Abstractor Architecture Details}
Each Abstractor $f_\ell$ is implemented as a two-headed Multi-Layer Perceptron with the following architecture:
\paragraph{Shared Backbone} A 3-layer feedforward network processes the input activation:
\begin{itemize}[noitemsep]
    \item Layer 1: Linear($d$, 1024) + LayerNorm + LeakyReLU(0.01) + Dropout(0.1)
    \item Layer 2: Linear(1024, 1024) + LayerNorm + LeakyReLU(0.01) + Dropout(0.1)
    \item Layer 3: Linear(1024, 1024) + LayerNorm + LeakyReLU(0.01)
\end{itemize}
\paragraph{Direction Head} Predicts the normalized direction vector:
\begin{itemize}[noitemsep]
    \item Linear(1024, 512) + ReLU
    \item Linear(512, $d$) + L2 Normalization
\end{itemize}
\paragraph{Magnitude Head} Predicts the scalar magnitude:
\begin{itemize}[noitemsep]
    \item Linear(1024, 512) + ReLU
    \item Linear(512, 1) + Softplus
\end{itemize}
The final prediction combines both: $\hat{a}_\ell(x) = \hat{m}_\ell \cdot \hat{d}_\ell$.
\paragraph{Training Configuration} Abstractors are trained using the AdamW optimiser with a learning rate of $5 \times 10^{-4}$ and a weight decay of $10^ {-3}$. We use batch size 128 with gradient clipping (max norm 1.0) and ReduceLROnPlateau scheduling (patience=10, factor=0.5). Training runs for up to 150 epochs with early stopping (patience=20). The contrastive margin threshold is set to 0.2, with loss weights $\lambda_{\text{repel}} = 0.75$ and $\lambda_{\text{mag}} = 1.0$.
\subsection{SFT Implementation Details}
\label{app:sft}
We employ PiSSA (Principal Singular values and Singular vectors Adaptation) for parameter-efficient fine-tuning, using the following configuration:
\paragraph{LoRA Parameters}
\begin{itemize}[noitemsep]
    \item Rank: $r = 16$
    \item Alpha: $\alpha = 16$
    \item Dropout: 0.05
    \item Target modules: query, key, value, and output projection layers
    \item Initialization: PiSSA (SVD-based initialization for faster convergence)
\end{itemize}
\paragraph{Training Data} We train on the English training set, using both content-laden and abstract examples. For $n$ content-abstract pairs in a fold, we create $2n$ training examples by including each variant separately. These examples are shuffled together during training, ensuring the model learns to handle both content-rich and symbolic inputs. This doubles the effective training set size compared to approaches that use only content-laden examples.
\paragraph{Training Hyperparameters} Learning rate and epochs vary by model to account for differences in architecture and baseline performance:
\begin{itemize}[noitemsep]
    \item Qwen-2.5-7B: lr=$10^{-5}$, 2 epochs
    \item Gemma-2-9B: lr=$10^{-5}$, 1 epoch
    \item Mistral-7B: lr=$10^{-5}$, 5 epochs
\end{itemize}

All models use batch size 4 with gradient accumulation over 2 steps (effective batch size 8). We apply gradient clipping (max norm 1.0) and early stopping based on validation loss (patience=10 epochs). The training objective is standard cross-entropy loss on ground truth labels.
\paragraph{Label Masking} To train the model to predict only the answer token, we mask all prompt tokens in the loss calculation. The system identifies the start of the answer by searching for "Valid" or "Invalid" tokens in the tokenised sequence, and then sets all preceding positions to -100 (an ignored index). If the answer position cannot be located, we conservatively mask the first 90\% of the sequence.
\subsection{CoT Implementation Details}
\label{app:cot}
Chain-of-Thought prompting encourages models to reason step-by-step before providing a final answer. We use the following prompt structure:
\begin{quote}
\small
\textit{"Evaluate the logical validity of the following syllogism by reasoning step-by-step. First, analyze each premise carefully. Then, determine whether the conclusion logically follows from the premises. Finally, state your answer as either 'valid' or 'invalid'. Ignore the meaning, realism, or plausibility of the statements. Let's think step by step (be concise and conclude within a few sentences):"}
\end{quote}
\paragraph{Generation Parameters} We use greedy decoding (temperature=0) with max\_new\_tokens=512 to allow sufficient space for reasoning. The system prompt varies by model family but generally presents the assistant as skilled in logical reasoning.
\paragraph{Answer Extraction} Since CoT responses contain both reasoning and the final answer, we extract the validity prediction using pattern matching. The system searches for explicit markers like "final answer", "therefore", or "the syllogism is" followed by "valid" or "invalid". If no clear marker exists, we examine the last 200 characters for standalone validity keywords.
\paragraph{Observed Failure Mode} Despite step-by-step reasoning, CoT frequently inherits the same belief bias present in direct answers. Consider this example where the model was asked to evaluate an invalid syllogism:
\begin{quote}
\small
\textbf{Premises:} All cows are mammals. Some mammals are not birds.\\
\textbf{Conclusion:} No birds are cows.\\
\textbf{Ground Truth:} Invalid

\textbf{Model Response:} \textit{"The given syllogism is a classic example of a valid deductive argument. [...] Since cows are mammals and there are mammals that are not birds, it is impossible for a bird to be a cow because a bird is not a mammal. Therefore, the given syllogism is valid."}

\textbf{Error:} The model incorrectly validates an invalid argument by confusing the logical structure. The premises do not support the conclusion: knowing that some mammals aren't birds tells us nothing about whether birds can be cows. The model's reasoning conflates semantic plausibility (birds obviously aren't cows in reality) with logical validity.
\end{quote}

This failure pattern--where the model generates seemingly logical reasoning that ultimately justifies a semantically plausible but logically unsupported conclusion (or vice versa)--appears consistently across belief-conflict cases, suggesting that CoT prompting alone does not resolve the underlying content effect.

\section{Dataset and Translation Details}
\label{app:dataset_details}
\subsection{Language Selection Rationale}
Our evaluation suite comprises ten languages selected to maximise typological and representational diversity across three key axes:

\paragraph{Family and Syntax} The languages span four major families: Indo-European (English, French, Spanish, Italian, German, Russian, Bengali), Sino-Tibetan (Chinese), Dravidian (Telugu), and Niger-Congo (Swahili). This selection includes varying syntactic orders--SVO (English, Chinese, Swahili) and SOV (Bengali, Telugu)--forcing the model to locate logical terms in different sequential positions. This tests whether the Abstractors target word order patterns or deeper structural abstractions.

\paragraph{Script and Tokenization} We include languages that challenge the tokenizer differently: shared Latin subwords (Romance/Germanic), distinct scripts (Cyrillic for Russian, Bengali script, Telugu script), and logographic systems (Chinese). If steering relied on token-level patterns rather than semantic abstractions, we would expect dramatic performance drops when script systems change. The robust transfer to Chinese and Russian (§\ref{subsec:crosslingual}) demonstrates that the intervention operates above the token level.

\paragraph{Resource Stratification} High-resource languages (French, Spanish, Italian, German, Russian, Chinese) receive substantial exposure during pre-training, resulting in well-structured latent representations. Low-resource languages (Swahili, Telugu, Bengali) receive limited exposure, yielding noisier representations. This stratification allows us to disentangle the quality of steering mechanisms from the quality of base-model representations: if steering fails on LRLs, is it because the mechanism itself is weak, or because the underlying representations are insufficient? Our results (§\ref{subsec:crosslingual}) support the latter interpretation.

\section{Design Rationale and Class-Specific Transformations}
\label{app:design_rationale}

\subsection{Unified vs. Conditional Architecture}
Preliminary analysis revealed that valid and invalid input syllogisms produce distinct activation patterns, suggesting that optimal transformations should be class-specific. While this observation motivates a conditional approach—using a classifier to gate separate Abstractors for each validity class—we opt for a \textbf{unified single-model architecture} in the main experiments for three reasons:

\paragraph{Error Propagation} Conditional systems introduce cascading failure modes: classifier errors compound with transformation errors, creating a failure ceiling determined by the weaker component.

\paragraph{Feasibility} Reliable validity probing, with performance matching the ones of the class-specific Abstractors, presupposes that the base model already encodes near-perfect internal representations distinguishing valid from invalid reasoning. However, if this were the case, it would be an important finding by itself. Our experiments confirm that validity probes achieve $\approx 95\%$ accuracy on English but degrade substantially in multilingual settings, limiting the practical effectiveness of conditional approaches.

\paragraph{Efficiency} Auxiliary classifiers add computational overhead and increase system complexity. Rather than relying on explicit gating, our unified Abstractor uses contrastive loss to \textit{implicitly} route activations based on their input geometry, automatically mapping valid and invalid inputs to their respective abstract manifolds.

\subsection{Oracle Validation: Class-Specific Upper Bound}
To validate that the limitations of our approach stem from routing rather than transformation quality, we implemented class-specific Abstractors. When applied in an "oracle" setting--where we use ground truth labels to select the corresponding valid or invalid Abstractor--we achieved \textbf{$\sim100\%$ accuracy across HRLs for all models}. This confirms that the observed failure modes arise solely from the difficulty of distinguishing valid from invalid logic paths.

To quantify the theoretical upper bound without an oracle, we trained validity probes on the last-token activations of each steering layer. We implemented this using a Support Vector Machine (SVM) with an RBF kernel, applied to standardized activation vectors with a stratified 80/20 train-test split. 

\begin{figure*}[h]
    \centering
    \includegraphics[width=1.6\columnwidth]{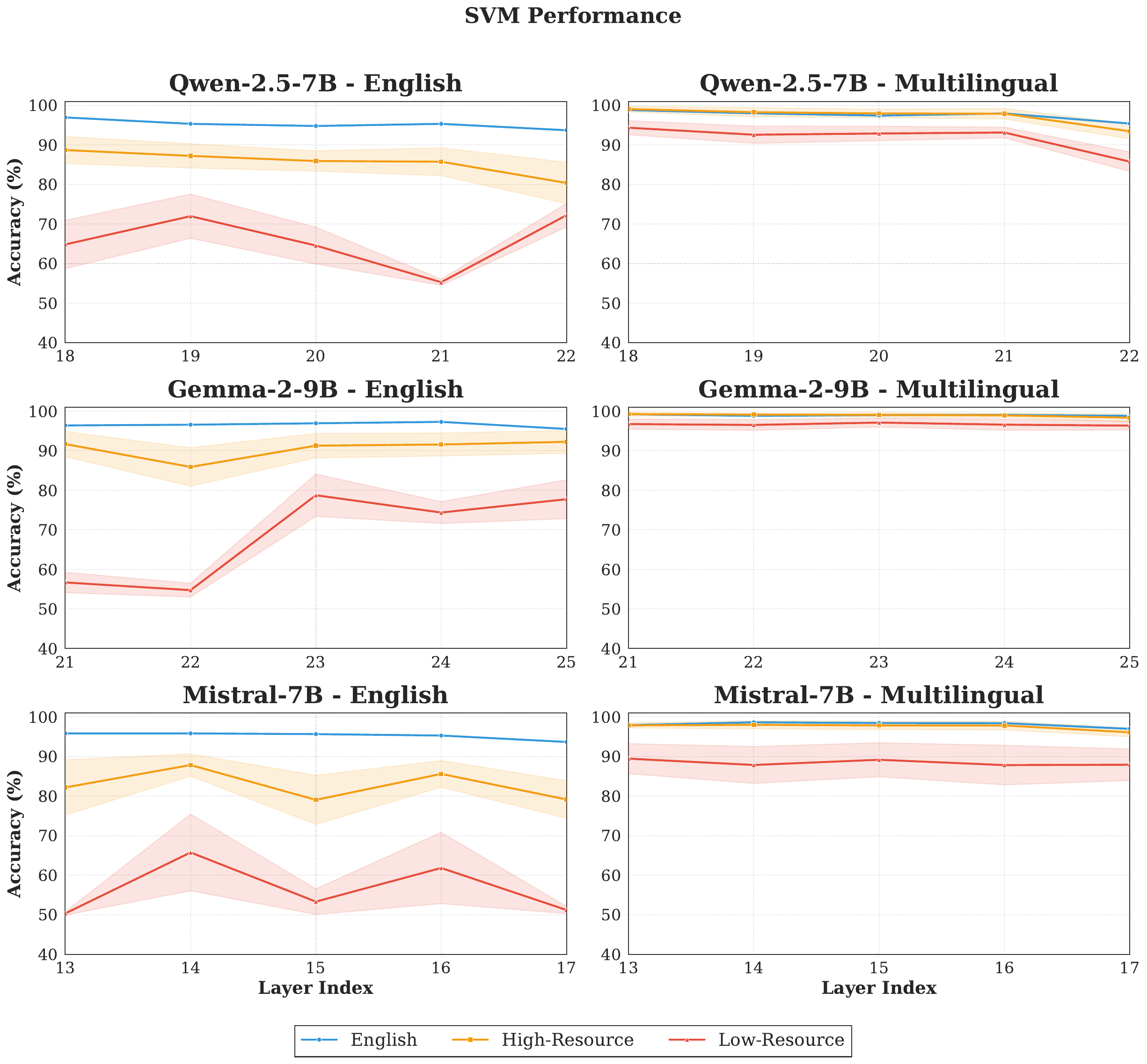}
    \caption{Validity probe (SVM with RBF kernel) accuracy across selected steering layers for \texttt{Qwen-2.5-7B}, \texttt{Gemma-2-9B} and \texttt{Mistral-7B}. The probe trained only on English maintains acceptable accuracy on HRLs but degrades severely on LRLs. Few-shot probes achieve optimal cross-lingual performance but undermine the zero-shot objective of our framework.}
    \label{fig:probe_accuracy}
\end{figure*}

Since the class-specific Abstractors achieve perfect accuracy when the ground truth is known, the bottleneck lies solely in the probe. Therefore, the results shown in Figure~\ref{fig:probe_accuracy} represent the \textit{hard upper bound} for a possible class-specific implementation. As observed, accuracy on non-English languages, particularly on LRLs, is only acceptable when the probe is trained on few-shot examples in the target languages. The English-trained probe exhibits a mild degradation on HRLs and a severe drop on LRLs, undermining the zero-shot generalisation objective of our design.

This analysis confirms our architectural choice: while class-specific transformations represent a theoretical upper bound, the unified approach is better suited for zero-shot generalisation, offering the best practical trade-off between performance, robustness, and system complexity.

\begin{table*}[h]
\section{Detailed Baseline Abstract Results}
\label{app:abstract_results}
\centering
\small
\resizebox{1.2\columnwidth}{!}{%
\begin{tabular}{l|cccc|cc}
\toprule
\multirow{2}{*}{\textbf{Model}} & \multicolumn{4}{c|}{\textbf{Categories}} & \multicolumn{2}{c}{\textbf{Global}} \\
\cmidrule(lr){2-5} \cmidrule(lr){6-7}
& \textbf{VP} & \textbf{VI} & \textbf{IP} & \textbf{II} & \textbf{Acc} & \textbf{BPA} \\
\midrule
\texttt{Qwen-2.5-7B} & 71.59 & 71.77 & 94.94 & 94.84 & 83.27 & 83.15 \\
\texttt{Qwen-3-14B} & 80.06 & 81.26 & 99.28 & 98.28 & 89.71 & 88.72 \\
\texttt{Gemma-2-9B} & 85.07 & 85.45 & 82.33 & 83.14 & 83.99 & 83.81 \\
\texttt{Gemma-3-12B} & 98.71 & 98.70 & 88.57 & 89.99 & 93.99 & 93.32 \\
\texttt{Mistral-7B} & 100.00 & 100.00 & 70.91 & 70.25 & 85.29 & 85.01 \\
\texttt{Ministral-14B} & 82.92 & 82.85 & 87.98 & 86.55 & 85.07 & 84.49 \\
\bottomrule
\end{tabular}%
}
\caption{Performance on abstract subset. VP/II are belief-consistent; VI/IP are belief-conflict. These results represent the upper bound that steering aims to approximate when processing content-laden inputs. The convergence of Accuracy (Acc) and BPA indicates a small content effect (content is removed), with minor variation attributable to statistical noise.}
\label{tab:abstract_results}
\end{table*}

\begin{figure*}[p]
\section{Steering Strength Ablation Study}
\label{app:strablation}
\subsection{Qwen-2.5-7B}
    \centering
    \includegraphics[width=\textwidth]{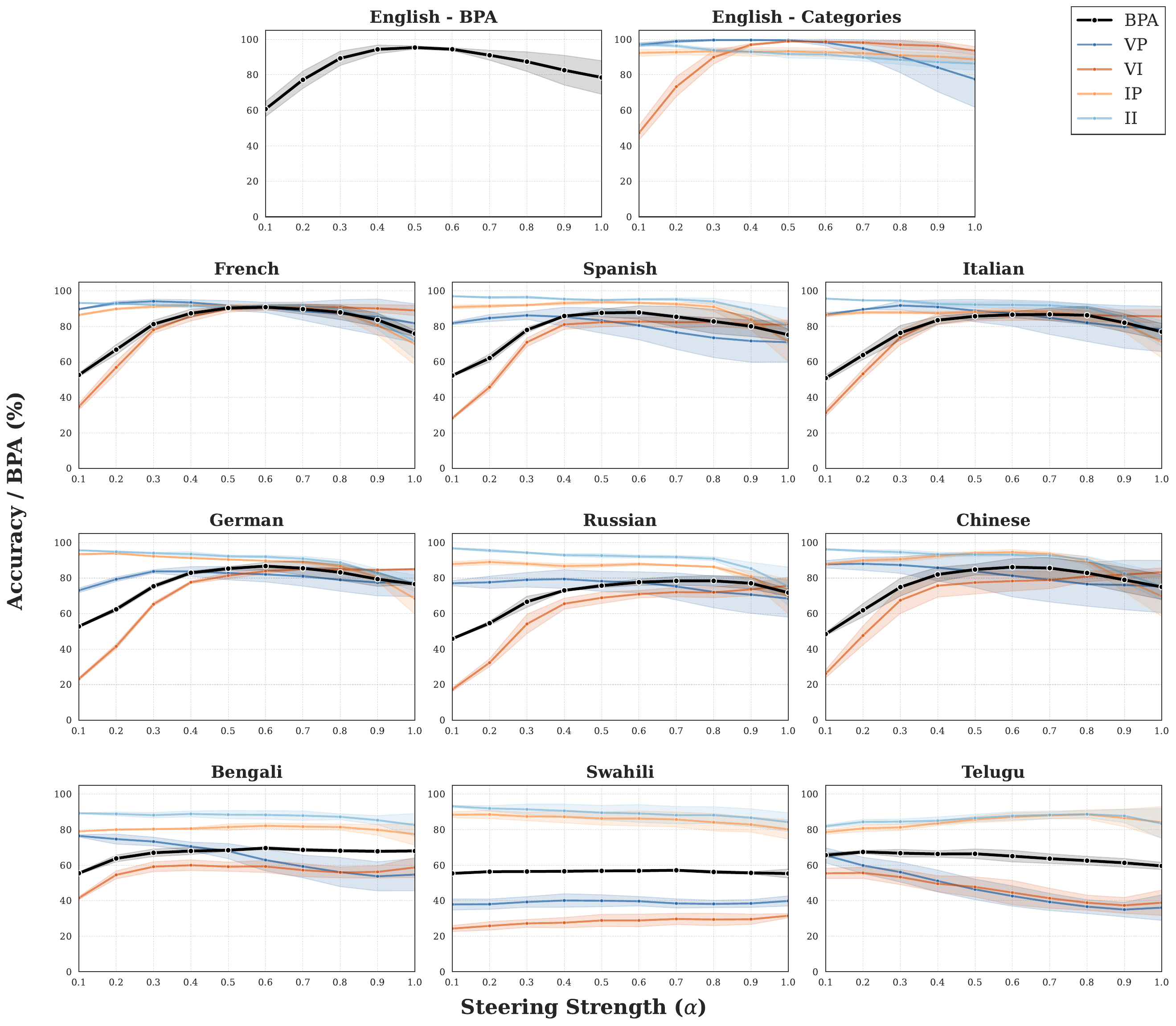}
    \caption{Steering strength ($\alpha$) ablation study for \texttt{Qwen-2.5-7B}, showing per-category accuracy and BPA (black) across all languages. VP/II (blue) are belief-consistent; VI/IP (orange) are belief-conflict.}
    \label{fig:qwen_ablation}
\end{figure*}
\begin{figure*}[p]
\subsection{Qwen-3-14B}
    \centering
    \includegraphics[width=1.0\textwidth]{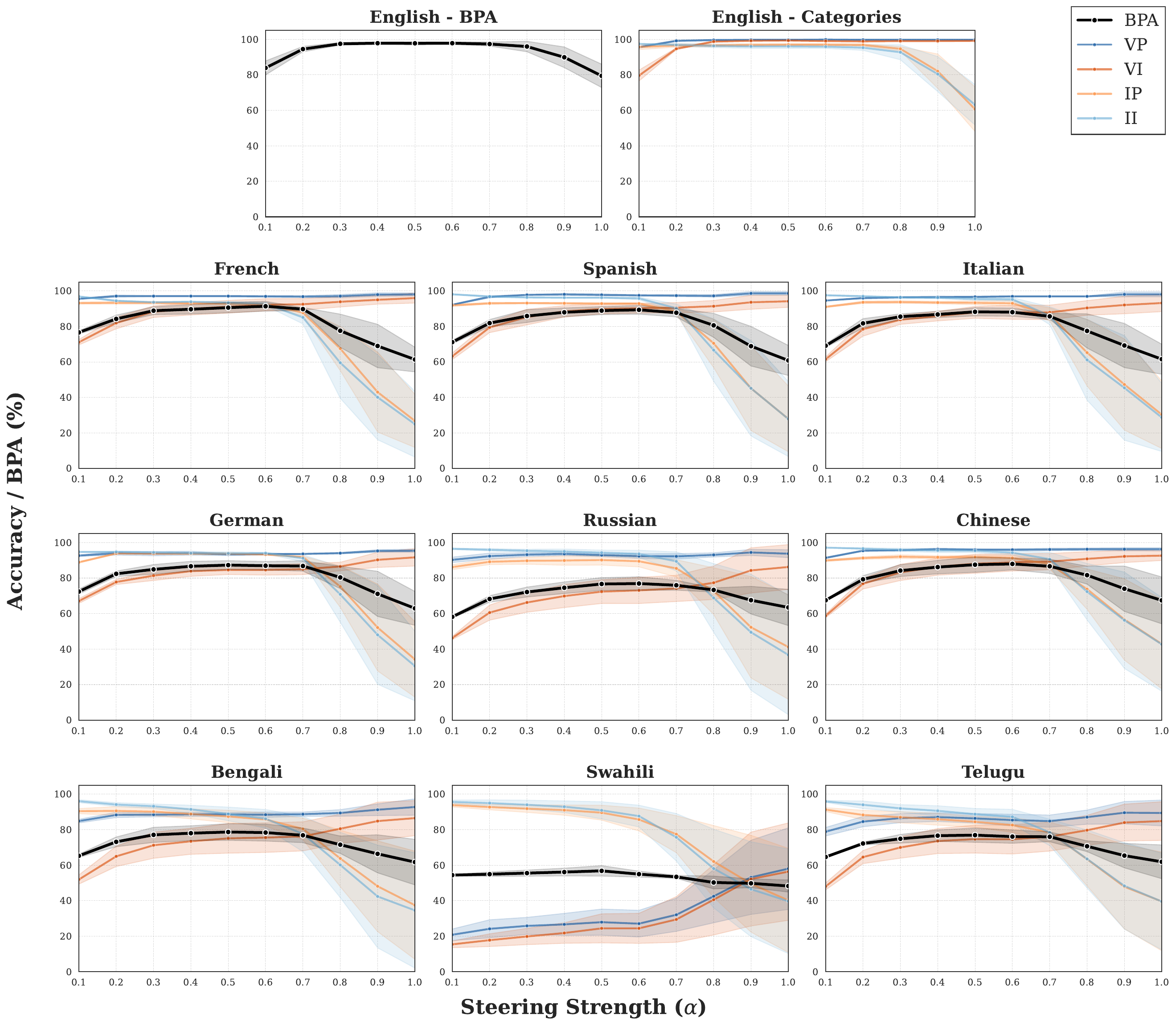}
    \caption{Steering strength ($\alpha$) ablation study for \texttt{Qwen-3-14B}, showing per-category accuracy and BPA (black) across all languages. VP/II (blue) are belief-consistent; VI/IP (orange) are belief-conflict.}
    \label{fig:}
\end{figure*}
\begin{figure*}[p]
\subsection{Gemma-2-9B}
    \centering
    \includegraphics[width=1.0\textwidth]{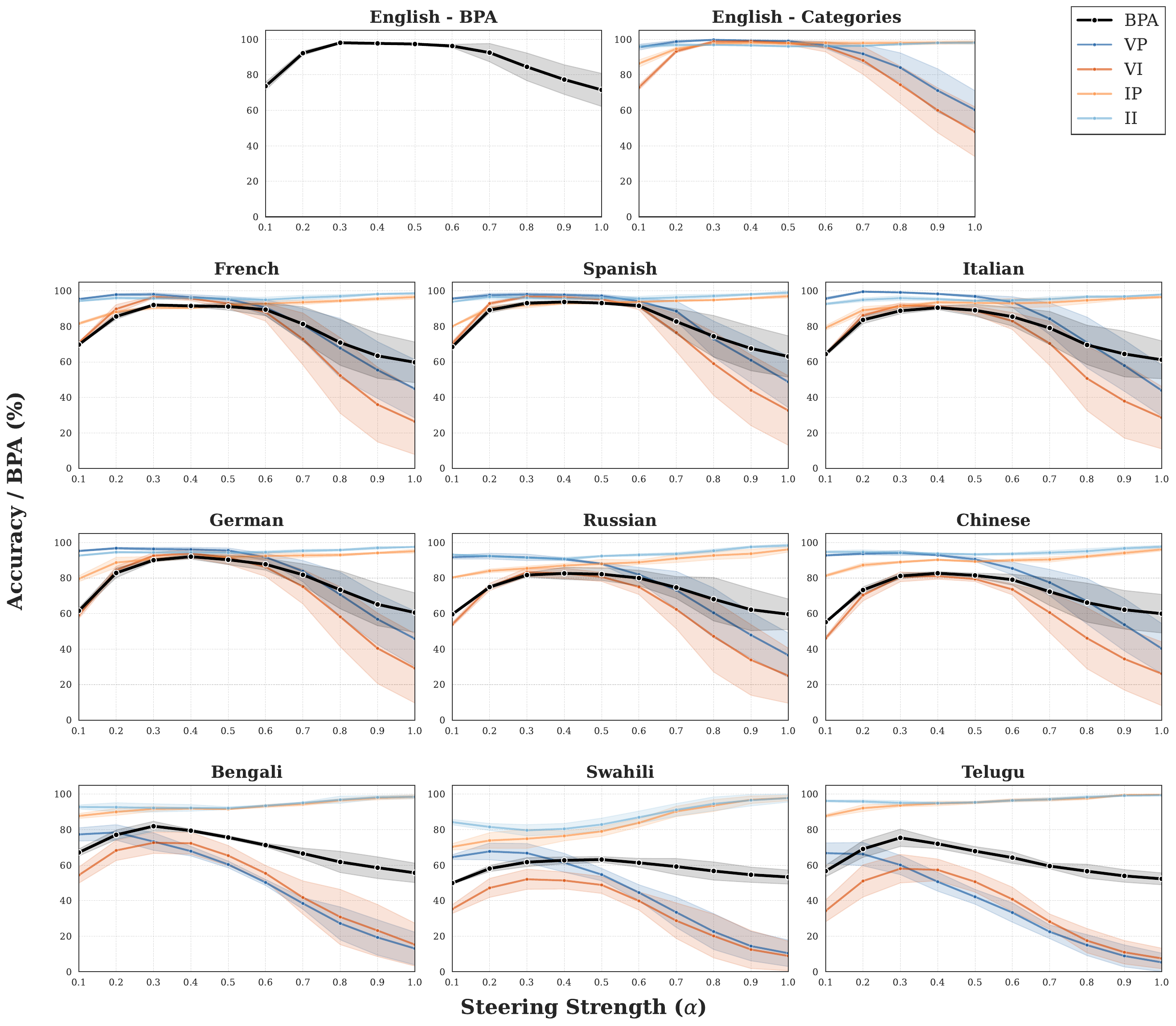}
    \caption{Steering strength ($\alpha$) ablation study for \texttt{Gemma-2-9B}, showing per-category accuracy and BPA (black) across all languages. VP/II (blue) are belief-consistent; VI/IP (orange) are belief-conflict.}
    \label{fig:}
\end{figure*}
\begin{figure*}[p]
\subsection{Gemma-3-12B}
    \centering
    \includegraphics[width=1.0\textwidth]{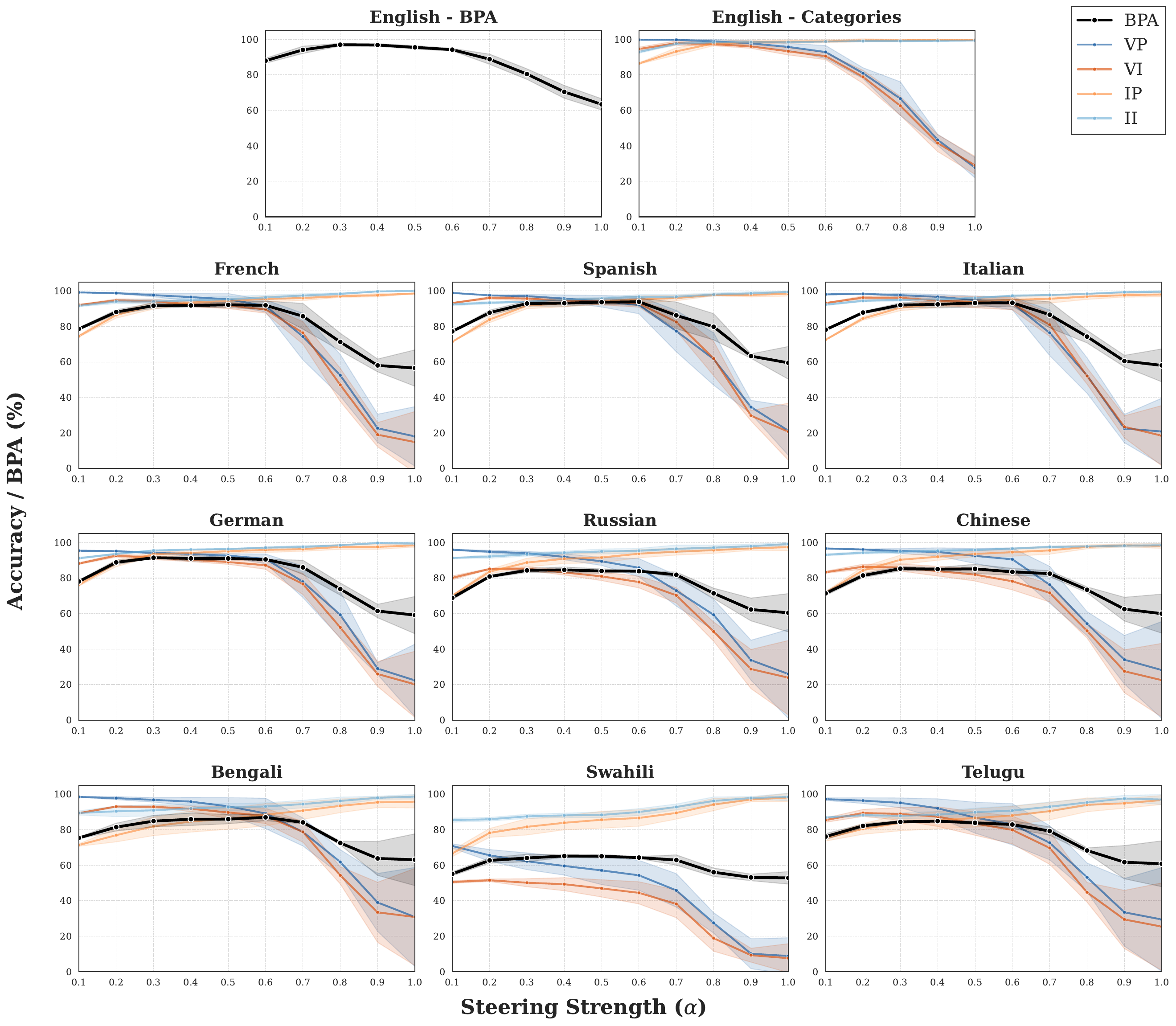}
    \caption{Steering strength ($\alpha$) ablation study for \texttt{Gemma-3-12B}, showing per-category accuracy and BPA (black) across all languages. VP/II (blue) are belief-consistent; VI/IP (orange) are belief-conflict.}
    \label{fig:}
\end{figure*}
\begin{figure*}[p]
\subsection{Mistral-7B}
    \centering
    \includegraphics[width=1.0\textwidth]{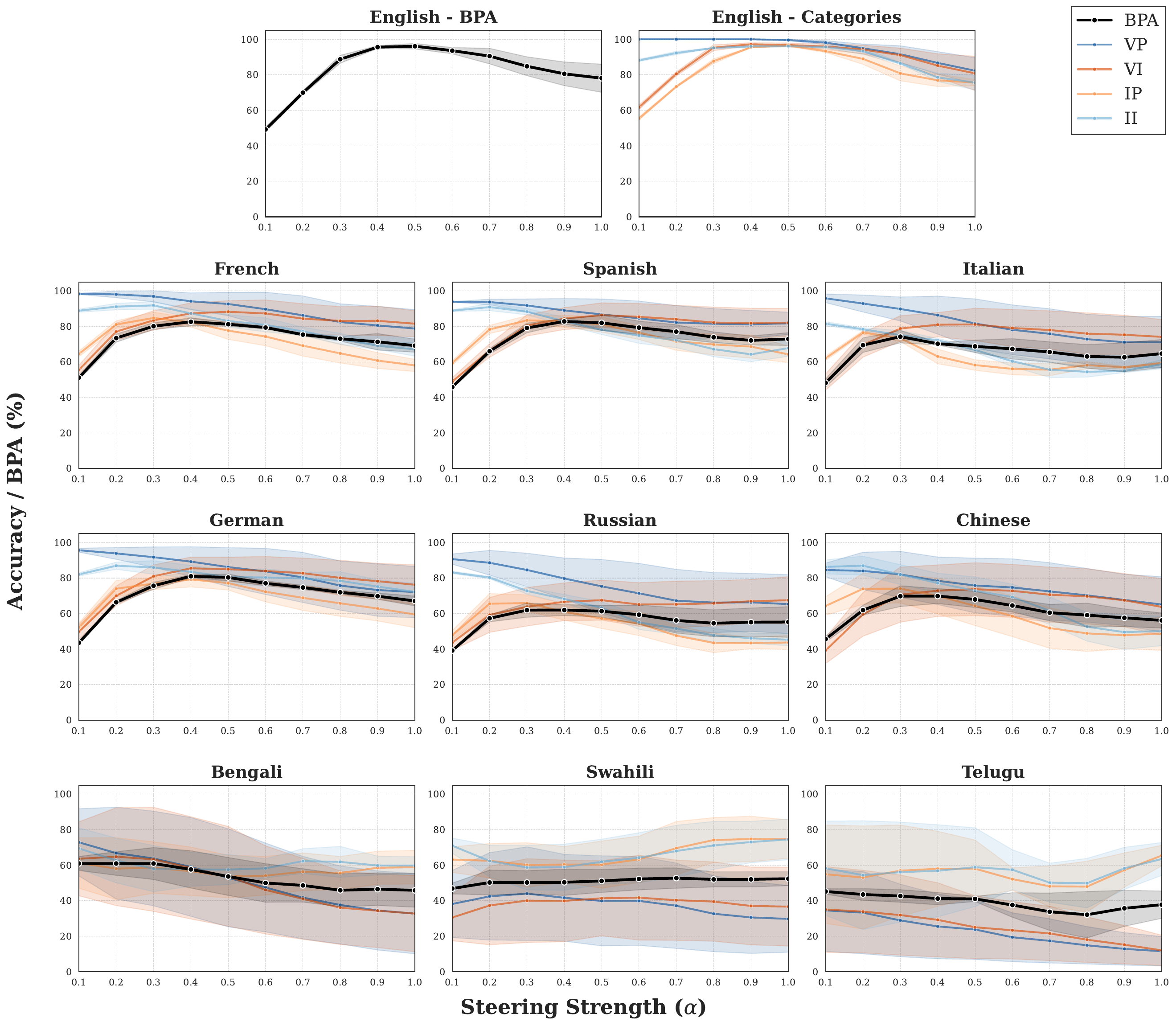}
    \caption{Steering strength ($\alpha$) ablation study for \texttt{Mistral-7B}, showing per-category accuracy and BPA (black) across all languages. VP/II (blue) are belief-consistent; VI/IP (orange) are belief-conflict.}
    \label{fig:}
\end{figure*}
\begin{figure*}[p]
\subsection{Ministral-14B}
    \centering
    \includegraphics[width=1.0\textwidth]{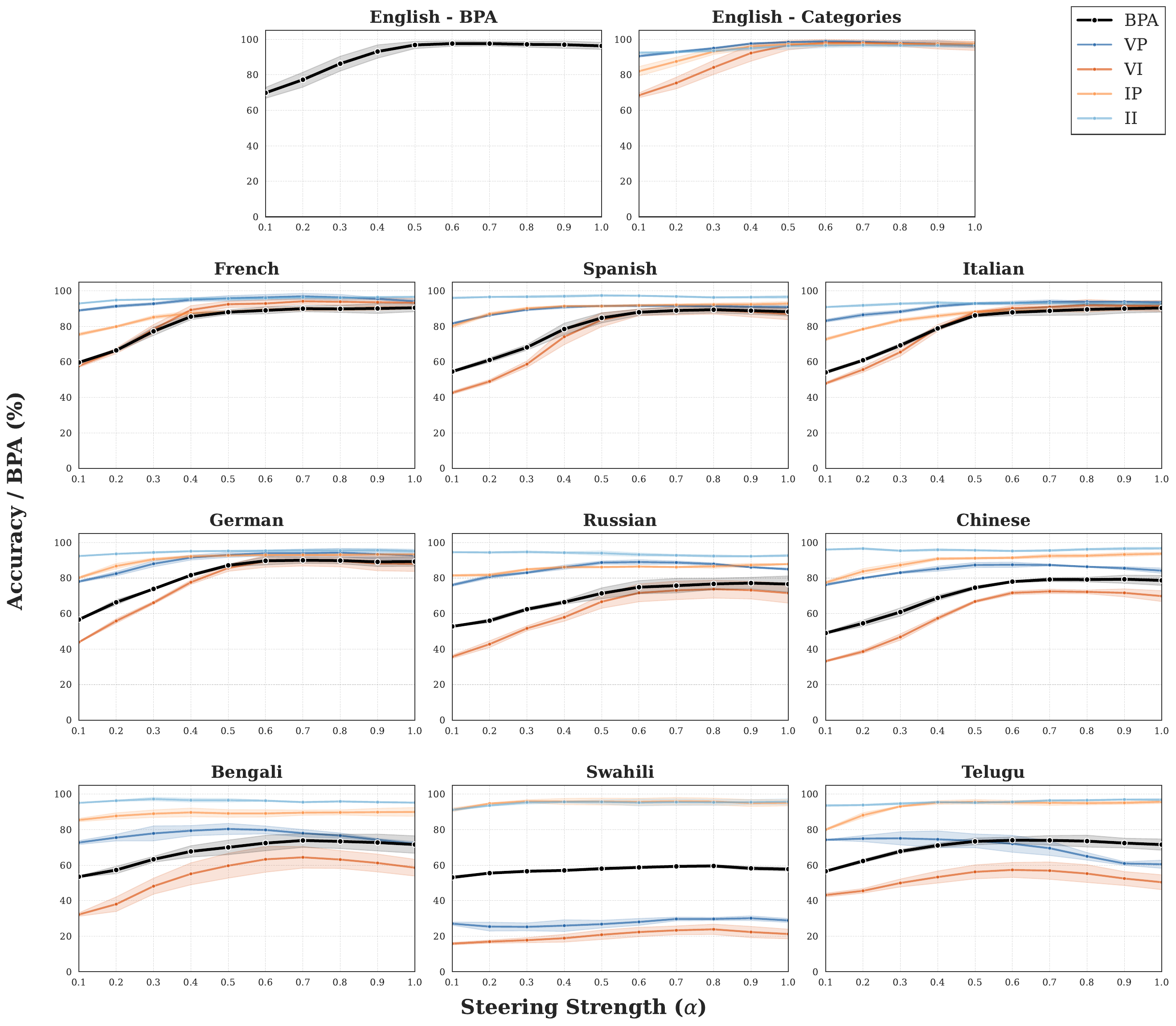}
    \caption{Steering strength ($\alpha$) ablation study for \texttt{Ministral-14B}, showing per-category accuracy and BPA (black) across all languages. VP/II (blue) are belief-consistent; VI/IP (orange) are belief-conflict.}
    \label{fig:}
\end{figure*}

\begin{table*}[t]
\section{Detailed Steering vs. SFT vs. CoT Results}
\label{app:steering vs sft vs cot}
\centering
\resizebox{0.95\textwidth}{!}{
\begin{tabular}{llccc}
\toprule
\textbf{Model} & \textbf{Method} & \textbf{English} & \textbf{HRLs} & \textbf{LRLs} \\
\midrule
\multirow{3}{*}{Qwen-2.5-7B} & Steering & 94.99 & 85.43 & 65.30 \\
 & SFT & 98.57 \textcolor{Green}{{\footnotesize(+3.77\%)}} & 84.63 \textcolor{Red}{{\footnotesize(-0.94\%)}} & 70.12 \textcolor{Green}{{\footnotesize(+7.38\%)}} \\
 & CoT & 81.65 \textcolor{Red}{{\footnotesize(-14.04\%)}} & 57.73 \textcolor{Red}{{\footnotesize(-32.43\%)}} & 55.00 \textcolor{Red}{{\footnotesize(-15.77\%)}} \\
\midrule
\multirow{3}{*}{Gemma-2-9B} & Steering & 98.25 & 89.65 & 76.62 \\
 & SFT & 98.78 \textcolor{Green}{{\footnotesize(+0.54\%)}} & 93.39 \textcolor{Green}{{\footnotesize(+4.17\%)}} & 76.96 \textcolor{Green}{{\footnotesize(+0.44\%)}} \\
 & CoT & 69.92 \textcolor{Red}{{\footnotesize(-28.83\%)}} & 58.54 \textcolor{Red}{{\footnotesize(-34.71\%)}} & 53.58 \textcolor{Red}{{\footnotesize(-30.08\%)}} \\
\midrule
\multirow{3}{*}{Mistral-7B} & Steering & 94.77 & 77.71 & 56.49 \\
 & SFT & 88.09 \textcolor{Red}{{\footnotesize(-7.05\%)}} & 78.17 \textcolor{Green}{{\footnotesize(+0.59\%)}} & 53.68 \textcolor{Red}{{\footnotesize(-4.98\%)}} \\
 & CoT & 34.25 \textcolor{Red}{{\footnotesize(-63.86\%)}} & 34.20 \textcolor{Red}{{\footnotesize(-55.99\%)}} & 49.97 \textcolor{Red}{{\footnotesize(-11.55\%)}} \\
\bottomrule
\end{tabular}%
}
\caption{Comparison of Bias-Penalized Accuracy (BPA) across Activation Steering, Supervised Fine-Tuning (SFT), and Chain-of-Thought (CoT) prompting. Percentage changes relative to steering are shown in green (improvement) or red (degradation).}
\label{tab:steering vs sft vs cot}
\end{table*}

\begin{figure*}[p]
\section{Detailed PPL Results}
\label{app:ppl}
    \centering
    \includegraphics[
        width=\textwidth,
        height=0.9\textheight,
        keepaspectratio
    ]{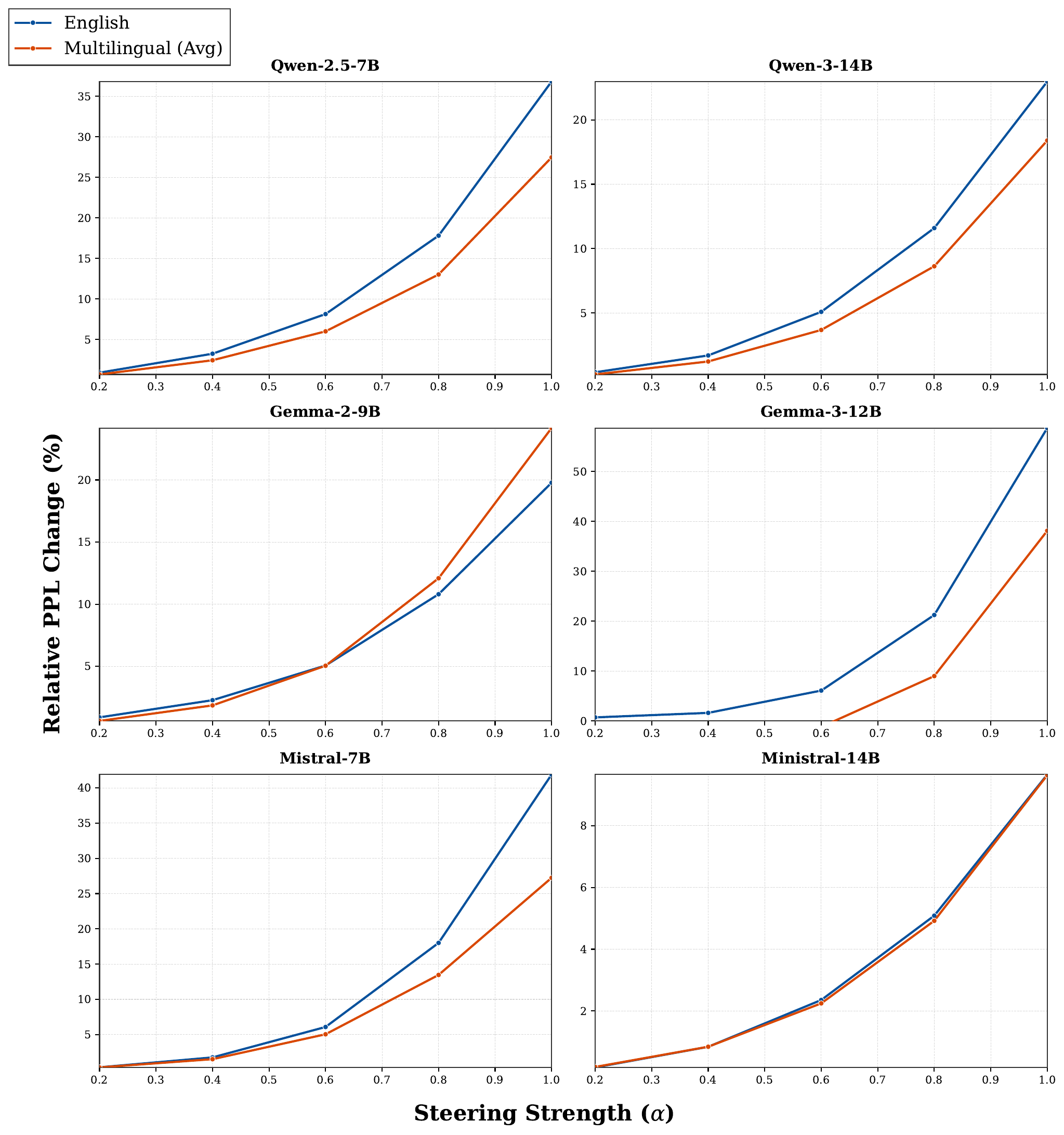}
    \caption{Perplexity degradation as a function of steering strength across all models and languages. At optimal steering configurations ($\alpha \approx 0.4-0.6$), fluency remains largely preserved. Significant degradation emerges only at aggressive steering levels ($\alpha \geq 0.8$).}
    \label{fig:}
\end{figure*}
\begin{figure*}[p]
\section{Detailed OOD Task Results}
\label{app:ood}
    \centering
    \includegraphics[
        width=\textwidth,
        height=0.9\textheight,
        keepaspectratio
    ]{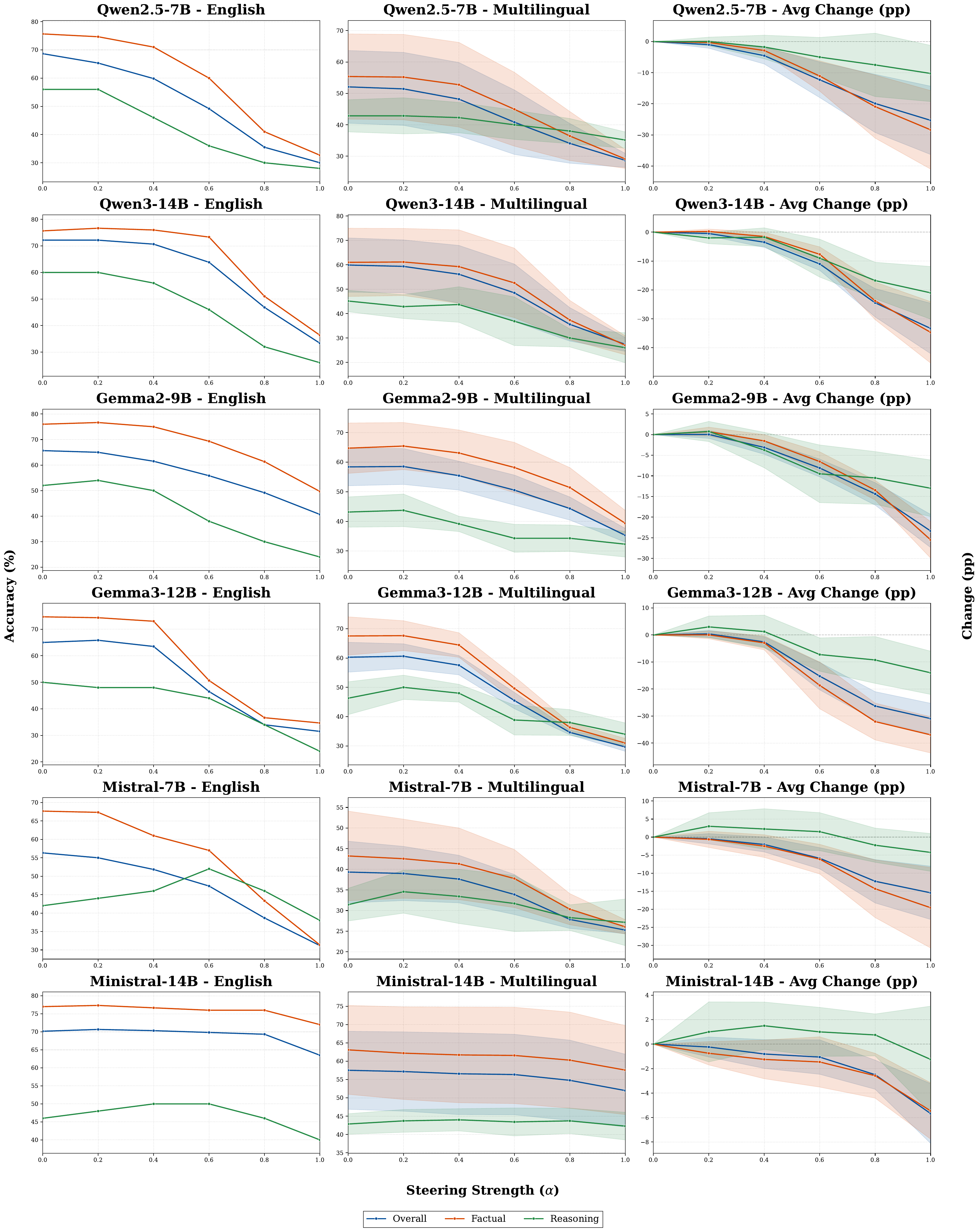}
    \caption{Performance impact on MMMLU subtasks across models. Factual knowledge tasks (orange) degrade sharply as steering suppresses semantic content, while Formal Logic tasks (green) often show more resilience. This divergence confirms the specificity of the intervention mechanism.}
    \label{fig:}
\end{figure*}

\end{document}